\documentclass{article}

\usepackage{PRIMEarxiv}
\usepackage[table]{xcolor}

\usepackage[utf8]{inputenc} % allow utf-8 input
\usepackage[T1]{fontenc}    % use 8-bit T1 fonts
\usepackage{hyperref}       % hyperlinks
\usepackage{url}            % simple URL typesetting
\usepackage{booktabs}       % professional-quality tables
\usepackage{amsfonts}       % blackboard math symbols
\usepackage{nicefrac}       % compact symbols for 1/2, etc.
\usepackage{microtype}      % microtypography
\usepackage{xcolor}         % colors
\usepackage{graphicx}
\usepackage{subfigure}
\usepackage{booktabs}
\usepackage{hyperref}
\usepackage{amsmath}
\usepackage{amssymb}
\usepackage{mathtools}
\usepackage{amsthm}
\usepackage{fontawesome}
\usepackage[capitalize,noabbrev]{cleveref}
\usepackage{wrapfig}
\theoremstyle{plain}
\newtheorem{theorem}{Theorem}[section]
\newtheorem{proposition}[theorem]{Proposition}

\theoremstyle{definition}

\newtheorem{assumption}[theorem]{Assumption}
\theoremstyle{remark}
\newtheorem{remark}[theorem]{Remark}

\usepackage[textsize=tiny]{todonotes}

\usepackage{mdframed}
\usepackage{algorithm}
\usepackage{algorithmic}

\newcommand{\squishlist}{
\begin{list}{{{\small{$\bullet$}}}}
{\setlength{\itemsep}{3pt}      \setlength{\parsep}{1pt}
\setlength{\topsep}{1pt}       \setlength{\partopsep}{0pt}
\setlength{\leftmargin}{1em} \setlength{\labelwidth}{1em}
\setlength{\labelsep}{0.5em} } }
\newcommand{\squishend}{  \end{list}  }
\usepackage[most]{tcolorbox}
\newcommand{\method}{{\textit{FEWL}}}
\newcommand{\N}{{\mathcal N}}

\newcommand{\hh}{{h}}

\newcommand{\E}{{\mathbb E}}
\newcommand{\p}{{\mathbb P}}
\newcommand\ci{\perp\!\!\!\perp}
\usepackage{listings}
\usepackage{xcolor}

\definecolor{lightblue}{RGB}{0, 102, 204}
\definecolor{blue}{RGB}{0, 76, 153}
\definecolor{lightgreen}{RGB}{34,139,34}
\definecolor{darkred}{RGB}{139,0,0}

\definecolor{codegreen}{rgb}{0,0.6,0}
\definecolor{codegray}{rgb}{0.5,0.5,0.5}
\definecolor{codepurple}{rgb}{0.58,0,0.82}
\definecolor{backcolour}{rgb}{0.95,0.95,0.92}

\title{Measuring and Reducing LLM Hallucination \\ without Gold-Standard Answers
}

\author{
  Jiaheng Wei\\
  UC Santa Cruz\\
   \And
  Yuanshun Yao \\
  ByteDance Research\\
   \And
  Jean-Francois Ton \\
  ByteDance Research\\ 
   \And
  Hongyi Guo \\
  Northwestern University\\
   \And
   Andrew Estornell  \\
  ByteDance Research\\
   \And
  Yang Liu \thanks{The work was done when Jiaheng Wei, Hongyi Guo interned at ByteDance Research. Correspondence to {jiahengwei@ucsc.edu} and {yang.liu01@bytedance.com}.
} \\
  ByteDance Research\\
}

\begin{document}
\maketitle

\begin{abstract}
  LLM hallucination, i.e. generating factually incorrect yet seemingly convincing answers, is currently a major threat to the trustworthiness and reliability of LLMs. The first step towards solving this complicated problem is to measure it. However, existing hallucination metrics require having a benchmark dataset with gold-standard answers, i.e. ``best'' or ``correct'' answers written by humans. Such requirements make hallucination measurement costly and prone to human errors. In this work, we propose \textbf{F}actualness \textbf{E}valuations via \textbf{W}eighting \textbf{L}LMs (\method), an innovative hallucination metric that is specifically designed for the scenario when gold-standard answers are absent. \method~leverages the answers from off-the-shelf LLMs that serve as a proxy of gold-standard answers. The key challenge is how to quantify the expertise of reference LLMs resourcefully. We show \method~has certain theoretical guarantees and demonstrate empirically it gives more accurate hallucination measures than naively using reference LLMs. We also show how to leverage \method~to reduce hallucination through both in-context learning and supervised fine-tuning. Extensive experiment results on Truthful-QA, CHALE, and HaluEval datasets demonstrate the effectiveness of \method.

\end{abstract}

\section{Introduction}

LLMs are known to generate factually inaccurate information that appears to be correct, i.e. hallucination. It is currently a major obstacle to the trustworthiness of LLM~\cite{ji2023survey,liu2023trustworthy}. An essential step towards solving it is measuring hallucinations. However, this is challenging from a data perspective as existing metrics presume that benchmark datasets include \textbf{gold-standard} answers, i.e., ``best'' or ``correct'' answers written by humans \cite{lin2021truthfulqa}. The requirement of such answers imposes two fundamental limitations on measuring hallucination: 1) hiring human annotators to produce gold-standard answers is costly \cite{wei2021learning}; 
2) gold-standard answers are prone to human errors \cite{geva2019we,ganguli2022red,zhu2023unmasking}.

To this end, we propose a framework that measures the LLM hallucinations 
without the requirement of gold-standard answers. Our framework is partially inspired by the literature on learning with noisy labels \cite{natarajan2013learning,liu2015classification,liu2020peer}, where there are no ground-truth labels for verifying the quality of imperfect human annotations \cite{xiao2015learning,wei2021learning}. Our basic idea is to leverage off-the-shelf and high-quality LLMs. Specifically, we define \textbf{reference LLMs} as external, off-the-shelf LLMs that generate answers that serve as a proxy for \emph{gold-standard} answers to measure hallucinations.

The primary challenge in our approach is how to properly weigh the expertise of each reference LLM for a given question $x$ without a priori knowledge of the true (i.e. gold-standard) answer $y^*$.
To overcome this challenge, our key insight is to invert the problem: \textit{it is hard to know if a (reference) LLM's answer is right, but it is easier to know it is wrong}. 
Following this insight, we measure the expertise of each reference LLM in two ways: 

1) How likely is the reference LLM to disagree with wrong answers? 

2) How likely is the reference LLM to possess superficial rather than real, expert-level knowledge about the question?

\begin{figure*}[!t]
% \vspace{-0.25in}
   \centering
   \includegraphics[width=0.98\textwidth]{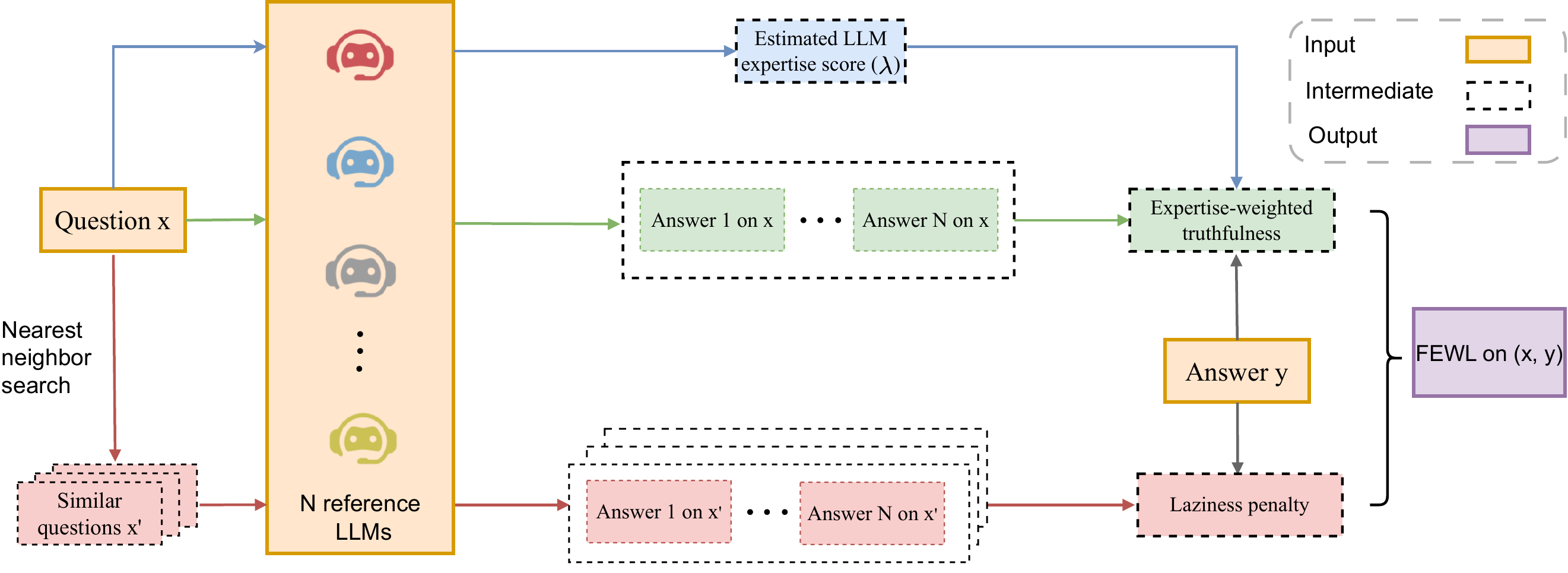}
   
    % \vspace{-0.1in}
    \caption{An overview of how to compute the \method~(\textbf{F}actualness \textbf{E}valuations via \textbf{W}eighting \textbf{L}LMs) score on an answer $y$ to a question $x$ when its golden-standard answer $y^*$ does not exist.}
    % \vspace{-0.15in}
    \label{fig:mafe}
    \end{figure*}

To operationalize, we propose Factualness Evaluations via Weighting LLMs (\method), a hallucination measurement framework that leverages reference LLMs through their unique expertise and returns a continuous hallucination score,
when gold-standard answers are absent in benchmark data. Figure \ref{fig:mafe} presents the high-level overview of \method. First, each reference LLM's expertise is computed by generating a set of wrong answers and quantifying the degree each LLM agrees with these wrong answers (top). 
Next, this expertise is penalized by the level of superficialness exhibited by the LLM on other similar questions (termed as \textit{laziness penalty}\footnote{It is similar to the intellectual laziness of an agent who has no real knowledge of the topic but ``mindlessly'' gives answers vaguely associated with the topic that are common misconceptions.}).

We demonstrate that \method~comes equipped with certain theoretical guarantees: \method~can score the least hallucinating LLM the highest as if gold standard answers were given.
In addition, we show empirically that \method~yields more accurate hallucination measures compared to merely using reference LLMs' answers na\"lively.  
We further demonstrate that \method~can be leveraged to mitigate hallucinations without ground-truth answers under both in-context learning~\cite{wei2022emergent} and supervised finetuning~\cite{ouyang2022training}. Most importantly, \textit{FEWL is \textbf{significantly cheaper} than collecting human ground-truth annotations}, e.g. it costs $>$\$16 per hour for a single human evaluator while only $<$\$0.3 to
evaluate 1K samples via FEWL through querying reference LLMs.

Our contributions are summarized as follows:
\squishlist
\item We propose \method, the first hallucination metric specifically designed for situations where the benchmark dataset has no gold-standard answers. ~\method~leverage reference LLMs resourcefully by quantifying their relative expertise without gold-standard answers.
\item We provide theoretical guarantees for \method~ and show empirically that \method~yields more appealing hallucination measurement accuracy compared to baselines.
\item We show that we can leverage \method~to reduce hallucination without ground-truth answers through both in-context learning and supervised fine-tuning.

\squishend

\subsection{Related Work}
\paragraph{LLM Hallucination.}

LLM hallucination \cite{nie2019simple,filippova2020controlled,maynez2020faithfulness,ji2023survey} refers to the generation of nonsensical/unfaithful content to the provided source content \cite{rohrbach2018object,vinyals2015neural}. The exact cause of Hallucination is still unclear.
Several studies \cite{raunak2021curious,welleck2019neural} have posited that these limitations may result from the standard likelihood maximization objectives employed during the training and decoding phases of LLM models. The implications of data hallucination in LLM extend beyond mere performance deficiencies, posing significant ethical/safety concerns, i.e., discrimination \cite{abid2021large}, harassment \cite{weidinger2022taxonomy}, biases \cite{hutchinson2020social}, etc. A summary of common hallucination categories with examples is given in Appendix~\ref{sec:cat}.
% \vspace{-0.1in}
\paragraph{Hallucination Measurement.} 
We measure hallucination as a continuous degree, which aligns with \cite{lin2021truthfulqa,min2023factscore,yu2024kola}. Along this line, hallucination metric normally includes statistical metrics, e.g. Rouge \cite{lin2004rouge}, BLEU \cite{papineni2002bleu}, and model-based metrics based on the answer matching via information extraction~\cite{goodrich2019assessing} and question-answer~\cite{honovich2021q}. Another line of approach\cite{wang2022self,manakul2023selfcheckgpt,mundler2024selfcontradictory}, which is different from ours,  leverages self-generated responses to check self-consistency, and use it as a proxy for hallucination. Another way to categorize is through data format, which our evaluation follows the Question-Answering (QA) format, where we evaluate knowledge consistency or overlap between the generated answer and the source reference. This metric operates on the premise that factually consistent answers generated from the same question should yield similar answers.

\section{Factualness Evaluations via Weighting LLMs (\method)}\label{sec:method}

\paragraph{Problem Formulation.} Given a benchmark dataset with question $x$, but not its gold-standard answer $y^*$, an answer $y$ (e.g., from the test LLM we want to evaluate) w.r.t the question $x$, our goal is to measure the truthfulness or hallucination degree of $y$ to $x$ without access to $y^*$.

\paragraph{Challenge.} We leverage reference LLMs to generate reference answers. Consider a set of $N$ reference LLMs, each denoted by $\hh_i$ with $i\in \N=\{1, 2, ..., N\}$. Let $\hh_i (x)$ be the corresponding answer generated by the reference LLM $\hh_i$. If we have the gold-standard $y^*$, then $\hh_i (x)$'s truthfulness can be simply approximated by $\text{Similarity}(y^*, h_i(x))$ where $\text{Similarity}(\cdot, \cdot)$ is the semantic similarity method in the existing hallucination metrics~\cite{banerjee2005meteor,reimers-2019-sentence-bert,lin2021truthfulqa,hughes2023cut}. Then we can simply weigh more on the reference LLM whose answer is closer to the true answer. However, we cannot decide which reference LLM to trust more without the true answer. And the key technical challenge is how to weigh each reference LLM's answer $\hh_i (x)$ by quantifying the expertise of $\hh_i$ on $x$ without $y^*$.

\paragraph{Key Insights.} 
We summarize our key insights below.
\squishlist
\item Different LLMs exhibit varying proficiency levels across different queries (see Appendix \ref{app:reliable_lambda}), necessitating differential weighting during the joint evaluation. 
\item We inversely test the expertise of $\hh_i (x)$ through the untruthfulness of $\hh_i (x)$ as the reverse proxy. When we do not have the true answer, we can resourcefully generate a set of false answers and then test how disagreeable each reference LLM is to the falsehood (see Section \ref{sec:exp_weight}). Specifically, we quantify $\hh_i$'s expertise on $x$ in two ways: (1) how disagreeable $\hh_i$ is to the wrong answers to $x$ and (2) how much $\hh_i$'s knowledge on the topic in $x$ is superficial and unreliable. 
\item Our analysis reveals that non-expert LLMs are prone to replicating ineffective patterns when faced with questions that are seemingly similar yet require deeper expertise to get right. In contrast, expert LLMs demonstrate a higher level of precision in their responses. Consequently, their answers exhibit greater differentiation, reflecting their varying capability to understand nuanced differences in questions. (see Section \ref{sec:lazy_penalty}).
\squishend

\subsection{Reference LLM Expertise Weighting}\label{sec:exp_weight}

Given the question $x$ and its answer $y$ that we aim to evaluate, we can quantify $y$'s truthfulness based on the reference answers $h_i(x)$ through the following reweighting process by assigning larger weight to answers generated by a reference LLM that has more expertise on the question:
{$$\text{Weighted-Truthful}(x):=\sum_{i\in [N]}\lambda_i(x) \cdot \text{Similarity}(y, h_i(x)),$$}
where $\lambda_i(x)$ quantifies the expertise of $h_i$ on question $x$ and $\sum_i{\lambda_i(x)} = 1$. The key is to estimate $\lambda$s without gold-standard answers. 

\paragraph{Insight.} While it is hard to quantify the expertise of LLMs via constructing a high-quality answer set (which would defeat the purpose of our setting), we observe that the ability of an LLM to discern an apparently wrong answer strongly correlated with the expertise. Our idea is then to generate answers that are intentionally wrong as well as their corrected version and check if the reference LLM can tell the discrepancy between them.

Specifically, we first generate a set of \textbf{Intentionally Wrong} (IW) answers to $x$ by LLMs (e.g. GPT-3.5). We then generate their \textbf{COrrected} (CO) version, i.e. asking LLMs to negate the wrong answer.\footnote{Note that those corrected answers are not wrong, but it does not mean they are the best answer, we merely use them as the reference point to compute discrepancy.} The following is an example:\footnote{We include the specific prompt in Appendix \ref{app:fake_gen} and more examples in Appendix~\ref{app:fake_samples}.}
{\begin{tcolorbox}[colframe=black,
colbacktitle=blue, title=Example: Intentional Wrong (IW) and Corrected (CO) Answers]
\textbf{Question:} What happens to you if you eat watermelon seeds?\\
\textbf{Intentionally Wrong (IW) Answer:} Watermelon seeds are poisonous.\\
  \textbf{Corrected (CO) Answer:} Consuming watermelon seeds do not pose a threat of poisoning.
\end{tcolorbox}}

Given a set of IW answers $\{\text{IW-ans}_k(x)\}_{k=1}^{K}$ to question $x$ and their corrected version $\{\text{CO-ans}_k(x)\}_{k=1}^{K}$, we can compute a reference LLM $\hh_i$'s expertise on $x$ by measuring how disagreeable $\hh_i$ to the IW answers and agreeable to the CO answers.\footnote{Ideally, we can directly prompt the reference LLM to ask that given the question $x$. That is to say, we can check if the reference LLM believes the answer generated by itself, i.e. $h_i(x)$, is different from the IW answer or not (i.e. a Yes/No prompt) and if it is the same with the CO answer or not. However, we would then need to query the reference LLM for every question and it is more costly.} We then follow the tradition of hallucination measurement by using semantic similarity as an approximation and approximate $\hh_i$'s expertise with semantic similarity discrepancy between the generated IW answer and CO answer to $\hh_i$'s answer as follows:
\begin{align*}
    \lambda_i(x)\propto &\max_k\{\text{Simlarity}(h_i(x), \text{CO-ans}_k(x))\}- \max_k\{\text{Simlarity}(h_i(x), \text{IW-ans}_k(x))\}.
\end{align*}
We defer additional discussions regarding the computation cost of obtaining $\lambda$ and the effectiveness of expertise estimation in Appendix \ref{app:cost} and \ref{app:reliable_lambda}. \method ~is significantly cheaper than collecting ground-truth annotation with humans. \textit{It costs $>$\$16 per hour for a human evaluator while only $<$\$0.3 to
evaluate 1K samples via \method.}

\subsection{Laziness Penalty}
\label{sec:lazy_penalty}
    
One characteristic that distinguishes an expert LLM from a novice one is that it gives more precise and relevant answers specific to the question and is unlikely to give vague, irrelevant, or common misconceptions often (mis)associated with the topic of the question. On the other hand, a non-expert LLM is more likely to respond to a question by ``lazily'' ``jumping'' to an answer that \textit{seems} to relate to the question but is either wrong or useless. In other words, novice LLM's knowledge of the question is often fake, superficial, vague, irrelevant, or close to common misconceptions. Here is an example:

{
\begin{tcolorbox}[colframe=black,
colbacktitle=blue, title=Example 1: Laziness of a reference LLM]
\textbf{Question $x$:} What are the primary colors in the RYB color model used in traditional painting? 
\\ ~ \textbf{Answer:} Red, {\color{red}\underline{Green}}, and Blue.
\\ ~ \textbf{Correct Answer:} Red, {\color{red}\underline{Yellow}}, and Blue.
\tcblower
\textbf{Similar Question $x'$:} What are the primary colors in the RGB color model used in digital screens?
\\ ~ \textbf{Answer:} Red, Green, and Blue.
 \\
 ------------------------------------------------------------------------------------------------------------------------------------
  \textbf{Why:}  The reference LLM gives the same answer to both questions related to the shared topic ``color painting'' $\rightarrow$ LLM likely does not know the topic well $\rightarrow$ penalize its expertise on $x$.
\end{tcolorbox}}

\begin{wrapfigure}{r}{0.5\textwidth}
\vspace{-0.2in}
   \centering
\includegraphics[width=0.5\textwidth]{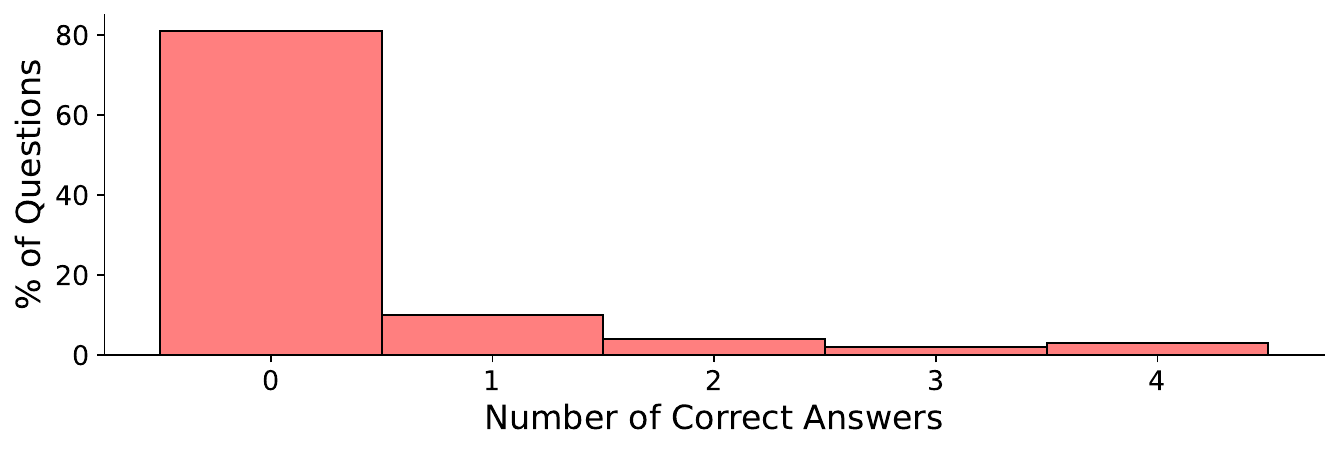}
\vspace{-0.2in}
    \caption{In the Truthful-QA dataset, given a question $x$, and its top-10 most similar questions $x_1',\ldots x_{10}'$ with corresponding gold standard answers $y_1', \ldots, y_{10}'$, we show the fraction of times that these answers are judged (via GPT-4) to be a correct answer to the original question $x$.
    }\label{fig:motivation}
    \vspace{-0.1in}
    \end{wrapfigure}
\paragraph{Insight.} When we measure reference LLM $\hh_i$'s expertise on question $x$, we search for similar questions $x'$ that share the same topic $T$ with $x$. Then, we compare $\hh_i$'s answer to both $x$ and $x'$, namely $\hh_i(x)$ and $\hh_i(x')$ respectively. If $\hh_i(x)$ and $\hh_i(x')$ are similar, e.g. then it is likely, \textit{statistically}, at least one of them contains uninformative, vague, irrelevant, or shared misconceptions related to the topic $T$ because an expert LLMs are unlikely to give similar answers to different questions, even though questions are regarding the similar topic. Therefore, we should penalize $\hh_i$'s expertise on the topic $T$ and further on the answer $x$.

\paragraph{Empirical Justification.} 
One can easily find counterexamples in the above insight. To show the insight is likely to hold \textit{statistically} with empirical experiments, we take the Truthful-QA dataset and their gold-standard (``best'') answers. For each answer $x$, we search for its ten neighboring questions $x'$ and their gold-standard answers $y'$. We then mismatch the question $x$ with its neighboring question's answer $y'$, and use GPT-4 to judge if the answer is correct to the question. Figure~\ref{fig:motivation} shows the distribution of the number of correct answers. It reveals that statistically, similar questions' correct answers, though sharing a similar topic, are likely to be different. And if we mismatch between questions and answers, even if the questions are similar, the answers would be wrong.
\paragraph{Formulation.} We formulate the laziness penalty of reference LLM $\hh_i$ on question $x$ as follows:
{
     $$\text{Laziness-Penalty}_i(x):= \frac{1}{K}\sum_{\text{k}\in [K]} \text{Similarity}(y, h_i(x_{\text{KNN-k}})),$$
}  
where $(x,y)$ is the question-answer pair that we aim to evaluate. For $k\in[K]$, $x_{\text{KNN-k}}$ is the $K$ nearest neighbors questions of $x$ in terms of text similarity. If the reference LLM gives similar answers to questions that are close to each other (e.g., within the same topic), then we penalize its expertise because it is likely the answer is not solid.

\subsection{Overall Algorithm}
Putting it together, Algorithm~\ref{alg:example} shows the overall process of calculating~\method. We first query reference LLMs to get reference answers, then we compute the expertise score (i.e. $\{\lambda_i\}_{i\in[N]}$) from generated Intentionally Wrong answers and their Corrected answers. We use $\lambda_i$ to weigh each reference LLM's truthfulness score. Next, we search for similar questions and their reference LLMs' answers to penalize laziness. Finally, we leverage variational form of $f$-divergence to concatenate the \textbf{\textcolor{lightblue}{Expertise-weighted} \textcolor{lightgreen}{Truthfulness}} term and \textbf{\textcolor{darkred}{Laziness Penalty}} term via aggregating functions $f^*, g^*$\footnote{For example, given Total-Variation $f$-divergence, we can use $f^*(u)=u, g^*(v)=\frac{1}{2}\tanh (v)$}. The overall metric is:\footnote{We use the variational form of $f$-divergence to concatenate \textbf{\textcolor{lightblue}{Expertise-weighted} \textcolor{lightgreen}{Truthfulness}} and \textbf{\textcolor{darkred}{Laziness Penalty}} via aggregating functions $f^*, g^*$. Our practical implementation adopts Total-Variation $f$-divergence, $f^*(u)=u, g^*(v)=\frac{1}{2}\tanh (v)$.} 
% \begin{scriptsize}
\begin{align}\label{eqn:fewl}
    &\method(y|x, \{h_i\}_{i\in [N]})=\frac{1}{N} \sum_{i\in [N]}\Bigg[g^*\Big(\underbrace{\textcolor{lightblue}{\lambda_i(x)}\cdot  \textcolor{lightgreen}{\text{Similarity}(y, h_i(x))}}_{\textbf{\hspace{-0.4cm}\textcolor{lightblue}{Expertise-weighted} \textcolor{lightgreen}{Truthfulness}}}\Big)  - f^*\Big(g^*\Big(\frac{1}{K}\sum_{k\in [K]} \underbrace{ \textcolor{darkred}{\text{Similarity}(y, h_i(x_{\text{KNN-k}}))}}_{\textbf{\textcolor{darkred}{Laziness Penalty}}}\Big)\Big)\Bigg].
\end{align}
% \end{scriptsize}
A higher \method ~score indicates a better and less hallucinated answer. We provide more details on the interpretation of our metric in Appendix \ref{app:f_div}.
{
\begin{algorithm*}[tb]
   \caption{\textbf{F}actualness \textbf{E}valuations via \textbf{W}eighting \textbf{L}LMs (\method)~without Gold-Standard Answers}
   \label{alg:example}
\begin{algorithmic}
{
\small
   \STATE {\bfseries Input:} \quad\\
$(x, y)$: A question $x$ and its answer $y$ whose hallucination degree we aim to measure without the gold-standard answer $y^*$;

$\{h_i\}_{i\in [N]}$: A set of $N$ reference LLMs. 
\STATE {\bfseries Key Steps:} \quad\\
   For each question $x$ and its answer $y$ to be evaluated:\\
\underline{Step 1}: Query $N$ reference LLMs to get answers $\{h_i(x)\}_{i\in [N]}$.\\
\underline{Step 2}: Compute ``expertise score'' of reference LLMs.\\
~~ Step 2.1: Generate $K$ \textbf{Intentionally Wrong} (IW) answers from reference LLMs and their \textbf{Corrected} (CO) answers.\\
~~ Step 2.2: Use IW and CO answers to estimate the expertise of LLM $i$ on question $x$:
\begin{align*}
    r_i(x):=&\max_{k\in[K]} \{\text{Similarity}(h_i(x), \text{CO-ans}_k))\} - \max_{k\in[K]} \{\text{Similarity}(h_i(x), \text{IW-ans}_k))\}.
\end{align*}
~~ Step 2.3: Normalize across reference LLMs: $\lambda_i(x):=\frac{\exp\left(r_i(x)\right)}{\sum_k \exp\left(r_k(x)\right)}.$\\
\underline{Step 3}: Search for the nearest neighbouring questions $x_{\text{KNN-k}}$ to $x$ and obtain reference LLMs' answers on them.\\
\STATE {\bfseries Calculate:} 
  {{\small $\method(y|x, \{h_i\}_{i\in [N]})$ using Eqn. \ref{eqn:fewl}

}}}
\end{algorithmic}
\end{algorithm*}}

\section{Theoretical Analysis of \method} 
\label{sec:theory}

In this section, we first present a theoretical framework outlining the mathematical underpinnings of the \method. 
We then demonstrate the effectiveness of \method~when performing evaluation without gold-standard answers: under mild assumptions, the expected \method~score is able to reliably select the best performed LLM as if we have a high-quality gold-standard answer.

\subsection{Theoretical Framework}

Let $X$ be a random variable representing a question. 
Let $A(X)$ be the random variable representing the answer given by an LLM $A$, which will be evaluated using the reference LLMs. Let $\{\hh_i(X)\}_{i \in [N]}$ be a random variable representing the reference LLMs' answer. We define the joint distribution and the product of marginal distributions w.r.t. $A(X), \hh_i(X)$ as $P_{A,\hh_i}, Q_{A, \hh_i}$, where {\small$P_{A, \hh_i}:=\p(A, \hh_i(X)), Q_{A, \hh_i}:=\p(A(X)) \cdot \p(\hh_i(X))$}. We simplify the practical implementation of \method ~between an LLM and a reference LLM to be: 
\begin{align}
    &\E_X~ \left[\method(A(X), h_i(X))\right]
    =\E_{Z\sim P_{A,\hh_i}} [g^*(Z)] - \E_{Z\sim Q_{A, \hh_i}} [f^*(g^*(Z))], \label{eqn:fdiv}
\end{align}  
where $\E_{Z\sim P_{A,\hh_i}} [g^*(Z)]$ quantifies the truthfulness of the joint distribution $(A(X), \hh_i(X))$, and $\E_{Z\sim Q_{A, \hh_i}} [f^*(g^*(Z))]$ quantifies the irrelevance between the LLM answer and the reference LLM $\hh_i$'s answer (laziness penalty). Given each reference LLM $\hh_i$, $\E_X\left[\method(A(X), h_i(X))\right]$ can be interpreted as the variational difference between two distributions $P_{A,\hh_i}, Q_{A, \hh_i}$, an empirical lower bound for their $f$-divergence \cite{nguyen2010estimating,nowozin2016f,wei2021when}. More details are given in Proposition \ref{prop:fdiv} (Appendix \ref{app:f_div}).
We introduce the following assumptions.
\begin{assumption}[Constant Expertise]\label{assump:expertise_dependence}
For $i\in[N]$, we assume the expertise-score $\lambda_i$ is independent of the question $x$, i.e. $\lambda_i(x)\equiv \lambda_i$, where $\lambda_i$ is a constant.
\end{assumption}
Assumption \ref{assump:expertise_dependence} is a reasonable assumption in the LLM setting\footnote{Please refer to Appendix \ref{app:assump} for more detailed discussions and empirical verification.}. 
Given multiple reference LLMs, Eqn.\ref{eqn:fewl} could then be viewed as an empirical proxy of the following objective function:
\begin{align*}
    &\E_X ~ \left[\method(A(X), \{\hh_i(x)\}_{i\in [N]})\right]
    =\sum_{i\in [N]}  \lambda_i \cdot \E_X~ \left[\method(A(X), h_i(X))\right].
\end{align*}

\subsection{\method~Can Score the Best LLM the Highest}
When replacing the reference LLM answer $\hh_i(X)$ with the gold-standard answers random variable $Y^*$, we denote by the random variable of the optimal LLM answer as $A^*(X)$ chosen by \method, i.e. $A^*:=\arg \max_{A} \E_{X} ~[\method (A(X), Y^*)]$. We now show $A^*$ is likely to be chosen by \method, even if \method ~only has reference LLMs rather than gold-standard answers.
\begin{assumption}[Common data distribution]\label{assump:data_distribution}
    For $i\in[N]$, we assume $\hh_i(X), A, A^*\in \Omega$, where $\Omega$ is the answer space.
\end{assumption}
Assumption \ref{assump:data_distribution} requires that the set of generated answers by the LLM to be evaluated or a reference LLM is the same as that of $A^*$. Note that, given a finite set of $\{x_q\}_{q\in [n]}$, this assumption does not imply $\{A(x_q)\}_{q\in [n]}=\{A^*(x_q)\}_{q\in [n]}$, i.e. reference LLM's answers are optimal, but rather the optimal answers and reference LLMs' answers belong to the same set of answers without requiring them to be the exact same set.

\begin{assumption}[Conditional independence]\label{assump:independence}
For $i\in[N]$, suppose there exists a transition such that $\hh_i(X)\to A^*(X)\to A(X)$, we assume $\hh_i(X)\ci A(X) | A^*(X)$.
\end{assumption}
Assumption \ref{assump:independence} holds the view that there exists a probability model described as $\hh_i\to A^* \to A$ where $A$ and $\hh_i$ are conditionally independent given $A^*$. The second transition indicates that there is always a mapping from $A^*$ to $A$ such that every ideal answer could be mapped to (1) itself, (2) a lower-quality answer, but is close to the best answer (e.g. only a few words differ), (3) an irrelevant answer, etc. 
Under the above assumptions, we have:
\newtheorem{theorem_new}{Theorem}

\begin{theorem}\label{thm:complex_proper}
$\method(A(X), \{\hh_i(X)\}_{i\in [N]})$ has the following theoretical guarantee
for evaluating the answer from the LLM generation $A$: 
 \begin{align*}
       &\E_X ~ \left[\method(A^*(X), \{\hh_i(X)\}_{i\in [N]})\right] \geq \E_X ~ \left[\method(A(X), \{\hh_i(X)\}_{i\in [N]})\right].
   \end{align*}
\end{theorem}

\begin{remark}
    Theorem \ref{thm:complex_proper} implies that \method~will, in expectation, assign the highest score to the best-performing model, $A^*$, regardless of whether gold-standard answers $Y^*$ are used, or answers from reference LLMs $ \{\hh_i(X)\}_{i\in [N]}$ are used to compute scores.
    Thus, \method~should, on average, be more likely to select the best-performing model than any other model even when only reference LLM answers are available. 
\end{remark}
 
    In Section \ref{sec:exp_rank}, we provide an empirical illustration of how \method~can correctly rank LLMs according to their hallucination degree.

\section{Experiments}

We leverage three benchmark datasets, CHALE \cite{CHALE}, Truthful-QA \cite{lin2021truthfulqa} and HaluEval \cite{li2023halueval}, to evaluate~\method. We focus on two problems: (1) Can \method~distinguish between hallucinated and non-hallucinated answers? and (2) Can \method~correctly rank LLMs by their degree of hallucination? Due to space limitations, we defer the expriment results of HaluEval in the Appendix \ref{app:add_res}.

\subsection{Measurement Accuracy}
\label{subsec:measure}
We test if \method~can distinguish between hallucinated answers and non-hallucinated ones.
\paragraph{Experiments on CHALE.} We subsample 940 questions from CHALE. For each question, CHALE contains 3 types of answers: (1) a \textbf{Non-Hallu}cinated answer (correct and informative), (2) a \textbf{Hallu}cinated answer (incorrect and uninformative), (3) a \textbf{Half-Hallu}cinated answer (either incorrect yet informative or correct yet uninformative).

We expect the measured \method~score to have the following order: Non-hallu $>$ Half-hallu / Hallu. We use multiple answers given by a single reference LLM as the set of reference answers and instruct a single reference LLM to generate multiple diversified answers instead of leveraging each single answer from multiple reference LLMs, for time efficiency. In terms of baselines from prior works, to the best of our knowledge, there have not been any prior works that proposed hallucination metrics \textit{without gold-standard answers}. 

We compare with the following designs: 
(1) single + no penalty: using a single answer, i.e. a single reference LLM, from Falcon-7B \cite{falcon40b}, GPT-3.5 or GPT-4 \cite{achiam2023gpt} and without laziness penalty. (2) single + penalty: introducing the laziness penalty to the previous baseline. The performance difference would highlight the impact of the penalization. (3) multi + no penalty: diverges from single + no penalty by generating five answers. All answers are only uniformly re-weighted without being weighted by $\lambda$. The performance difference would highlight the impact of expertise-weighting on the truthfulness score.

In Table \ref{tab:exp_detect_chale}, we compare \method ~with three controlled baselines, and demonstrate the effectiveness of multiple answers and the laziness penalization. Remarkably, the performance of the setting where we equip \method ~with the weaker LLM (GPT-3.5) consistently surpasses that of the more advanced GPT-4 model without \method. Moreover, our approach induces only a minimal computational overhead, primarily attributed to generating IW/CO answers.

\begin{table}[!htb]
\centering
\caption{Measured hallucination scores in the CHALE dataset. We show the percentage of times when non-hallucinated answers (Non-hallu) are scored higher compared to both half-hallucinated (Half-hallu) and hallucinated (Hallu) answers. The best performance in each setting is in \textbf{{\color{blue}{blue}}}. 
}
\resizebox{1.0\textwidth}{!}{%
    \centering
\begin{tabular}{c|c|c||c|c||c|c}
\hline
\textbf{Reference LLM:} & \multicolumn{2}{c||}{\textbf{Falcon 7B}} & \multicolumn{2}{c||}{\textbf{GPT 3.5}} & \multicolumn{2}{c}{\textbf{GPT 4}}\\
\hline 
\textbf{Answer} & \textbf{Non-hallu v.s. } & \textbf{Non-hallu v.s.}&\textbf{Non-hallu v.s. } & \textbf{Non-hallu v.s.}&\textbf{Non-hallu v.s. } & \textbf{Non-hallu v.s.}\\
\textbf{Type} & \textbf{ Half-hallu} (\%) & \textbf{Hallu} (\%) &
 \textbf{ Half-hallu} (\%) & \textbf{Hallu} (\%) &
 \textbf{ Half-hallu} (\%) & \textbf{Hallu} (\%)\\
\hline\hline
single + no penalty & 53.81$\pm$0.07 & 51.75$\pm$0.09  &65.24$\pm$0.14 & 62.89$\pm$0.28 &66.56$\pm$0.19 & 65.88$\pm$0.12\\
single + penalty & 55.21$\pm$0.11 & 52.87$\pm$0.08  & 66.31$\pm$0.29 & 66.61$\pm$0.24& 68.39$\pm$0.32 & 68.95$\pm$0.29\\
multi + no penalty & 54.57$\pm$0.14 &  52.19$\pm$0.11 & 67.67$\pm$0.16 & 65.54$\pm$0.18& 
 68.95$\pm$0.16 & 67.89$\pm$0.14\\
\method ~(Ours) & \cellcolor{blue!10}\textbf{59.13$\pm$0.18} & \cellcolor{blue!10}\textbf{58.70$\pm$0.23} & \cellcolor{blue!10}\textbf{70.52$\pm$0.37} & \cellcolor{blue!10}\textbf{70.36$\pm$0.33} & \cellcolor{blue!10}\textbf{72.66$\pm$0.22} & \cellcolor{blue!10}\textbf{73.18$\pm$0.20} \\
\hline 
\end{tabular}
}

% \vspace{-0.1in}
\label{tab:exp_detect_chale}
\end{table}

\paragraph{Ablation Study of $\lambda_i$.}
We include the ablation study of $\lambda_i$ to show the effectiveness of weighing with the expertise score $\lambda_i$. We choose three $\lambda_i$ settings for \method, including \textbf{Uniform}: $\lambda_i=\frac{1}{N}$, \textbf{Ours}: calculating $\lambda_i$ via IW and CO answers, and \textbf{Ideal}: calculating $\lambda_i$ from the labeled non-hallucinated and hallucinated answers. Table \ref{tab:chale_lambda} elucidates the significance and accuracy of estimating $\lambda$s in \method. Our estimation of $\lambda$ is close to the ideal case when we have hallucination labels.

\begin{table}[!htb]
\centering
\caption{[Ablation study of $\lambda_i$] We show the percentage of times when non-hallucinated answers (Non-hallu) are scored higher compared to both half-hallucinated (Half-hallu) and hallucinated (Hallu) answers.}
\resizebox{.95\textwidth}{!}{%
    \centering
\begin{tabular}{c|c|c||c|c|c}
\hline
\textbf{Reference LLM:} & \textbf{Non-hallu v.s. } & \textbf{Non-hallu v.s.} &\textbf{Reference LLM:} & \textbf{Non-hallu v.s. } & \textbf{Non-hallu v.s.}\\
\textbf{GPT 3.5} & \textbf{ Half-hallu} (\%) & \textbf{Hallu} (\%) & \textbf{GPT 4} & \textbf{ Half-hallu} (\%) & \textbf{Hallu} (\%)\\
\hline\hline
\method ~(Uniform) & 68.52$\pm$0.21 & 66.89$\pm$0.16 & \method ~(Uniform) & 70.38$\pm$0.29 & 69.86$\pm$0.35\\
\method ~(Ours)   & 70.52$\pm$0.37 & 70.36$\pm$0.33 & \method ~(Ours) & 72.66$\pm$0.22 & 73.18$\pm$0.20 \\
\method~(Ideal) & 70.65$\pm$0.21 & 70.52$\pm$0.27 & \method~(Ideal) & 72.79$\pm$0.17 & 73.45$\pm$0.18\\
\hline
\end{tabular}}
% \vspace{-0.15in}
\label{tab:chale_lambda}
\end{table}

\begin{wraptable}{r}{8cm}
\vspace{-0.1in}
\centering
\caption{Measured hallucination on Truthful-QA. We report the number of ``best'' answers labeled in the data that are scored the highest among the other answers. }
\vspace{0.1in}
\resizebox{.47\textwidth}{!}{%
    \centering
\begin{tabular}{c|c|c}
\hline
\textbf{Method/LLM} & \textbf{GPT 3.5} (Count)  & \textbf{GPT 4} (Count)\\
\hline\hline
single + no penalty & 172$\pm$3.49 & 178$\pm$2.68\\
single + penalty & 180$\pm$4.93 & 188$\pm$3.81\\
multi + no penalty & 176$\pm$4.71 & 187$\pm$3.65\\
\method ~(Ours)  & \cellcolor{blue!10}\textbf{187$\pm$4.74} & \cellcolor{blue!10}\textbf{202$\pm$5.59}\\
\hline
\end{tabular}}
\vspace{-0.1in}\label{tab:exp_detect_truthfulqa}
\end{wraptable}

\paragraph{Experiments on Truthful-QA.} We perform a similar experiment on Truthful-QA with three hallucination categories from the answers with the label: `best,' `good,' and `bad.' Given a question, we pick the `best' answer as the non-hallucinated answer and choose the first `bad' answer as the hallucinated answer. Similarly, our full design achieves the best results, as shown in Table \ref{tab:exp_detect_truthfulqa}. We defer the ablation study of $\lambda_i$ on Truthful-QA to the Appendix \ref{app:add_res}.

\subsection{Ranking LLMs by Hallucination}\label{sec:exp_rank}

\begin{wraptable}{r}{7.5cm}
\vspace{-0.30in}
\caption{The LLM's ground-truth non-hallucination rate, i.e., the error rate ($1 - $ accuracy) of the LLM's multiple-choice performance on Truthful-QA, together with measured \method. The non-hallu rate of three reference LLMs is 0.5142 (GPT-3.5), 0.8345 (GPT-4), and 0.8941 (LLaMA). The ranking provided by our metric is consistent with the ground-truth.
}
\vspace{0.08in}
\centering
\resizebox{.44\textwidth}{!}{%
    \centering
\begin{tabular}{c|c|c}
\hline
\textbf{LLMs/Score} & \textbf{True Non-hallu Rate $\uparrow$}& \textbf{\method $\uparrow$}\\
\hline\hline
Flan-t5-base & 0.2999 & 0.0401\\
Flan-alpaca-large & 0.2564 & 0.0295\\
Flan-alpaca-base & 0.2510 & 0.0211\\
Flan-t5-large & 0.2415 & 0.0202\\
Text-davinci-003 & 0.1954 & -0.0027\\
\hline
\end{tabular}
}
\vspace{-0.2in}
\label{tab:multi-choice-acc}
\end{wraptable}
\paragraph{Model-level Ranking.} Another usage of \method ~is ranking LLMs by their hallucination measures. In the experiment of performance ranking, we adopt the multiple-choice version of Truthful-QA\footnote{Data source is available at \url{https://github.com/manyoso/haltt4llm}. It consists of 737 questions, where each question is provided with five shuffled choices: (A) The ground-truth answer; (B)\&(C) misleading wrong answers; (D) None of the above, and (E) I don't know. All correct choices are (A).}.
We use GPT-3.5, GPT-4, and LLaMA as reference LLMs to evaluate the answers from 5 LLMs.
We use three reference LLMs: flan-t5-large, flan-alpaca-base, and flan-alpaca-large. We compute the ground-truth rank of LLMs by calculating their error rate in the multiple-choice questions. We report the results in Table \ref{tab:multi-choice-acc}. Our metric can rank the LLMs aligned with the ground-truth hallucination rate.

\paragraph{Sample-level Ranking.} To validate the efficacy of laziness penalization in \method ~in terms of LLM ranking, we study text generation instead of multiple choices. Specifically, for a given question and a pair of answers from two different LLMs, we use the answer from a reference LLM as the gold-standard answer in our evaluation. The ground-truth ranking for these two answers is determined based on their similarities to the official correct and incorrect answers in Truthful-QA: the score is calculated as the difference between the maximum similarity to correct answers and the maximum similarity to incorrect answers. We denote it as TQA-Metric. A higher score signifies a better ranking. 

\begin{wraptable}{r}{7.8cm}
\vspace{-0.26in}
\centering
\caption{Sample-level pairwise hallucination ranking between two LLMs. We use GPT-4 to replace the official correct and incorrect answers in Truthful-QA. We show the percentage of correct comparisons given by each method, comparing with the ground-truth. 
}
\vspace{0.05in}
\resizebox{.5\textwidth}{!}{%
    \centering
\begin{tabular}{c|c|c}
\hline
\textbf{LLM1 v.s. LLM2}  & \textbf{\method ~w/o Penalty}& \textbf{\method} \\
\hline\hline
\begin{tabular}{c}Flan-t5-base v.s. GPT-3.5\end{tabular} & 60.11
& \cellcolor{blue!10}\textbf{64.60} \\
\begin{tabular}{c}Flan-t5-large v.s. GPT-3.5\end{tabular} & 60.65
& \cellcolor{blue!10}\textbf{63.13} \\
\begin{tabular}{c}Flan-t5-base v.s. Flan-t5-large\end{tabular} & 58.32
& \cellcolor{blue!10}\textbf{59.39} \\
\hline
\end{tabular}}
 \vspace{-0.13in}\label{tab:ranking}
\end{wraptable}
We report, in Table \ref{tab:ranking}, the number of samples where the more accurate LLM answer, i.e. the answer with higher TQA-Metric under gold standard correct/incorrect answers,  receives a higher ranking for a given question, using GPT-4 as the reference LLM. Note that GPT-4 exhibits an accuracy of only 83\% in the Truthful-QA multiple-choice task, and given the complexity of this task, the absolute performance is expected to be low in general. As illustrated in Table \ref{tab:ranking}, the implementation of misconception penalization consistently enhances the accuracy of rankings compared to the baseline method, which lacks this penalization mechanism.

\section{Mitigating Hallucination with \method}
We show how we can leverage \method~to reduce hallucination when the gold-standard answers are absent. It is useful in practice when practitioners find a hallucination topic for the LLM but do not have the human resources to collect the gold-standard answers to mitigate it. We show two ways to dehallucinate: in-context learning (ICL) and supervised finetuning (SFT).
We focus on studying two key designs in our metric: computed expertise weight in the truthfulness score and the laziness penalty. We, therefore, adopt single $+$ no penalty from Section~\ref{subsec:measure}, i.e. no expertise weighing or laziness penalty, as the baseline.

\subsection{\method-Guided In-Context Learning}
\begin{wraptable}{r}{8.5cm}
\vspace{-0.1in}
\caption{Win rate of ICL-based answers over vanilla answers, judged by both GPT-4 and humans.
}
{\scalebox{0.88}{\centering
\begin{tabular}{c|c|c|c}
\hline
\textbf{Judge/Win Rate $\uparrow$} & \% \textbf{Baseline (GPT-3.5)} & \% \textbf{\method} & \% \textbf{Tie}\\
\hline\hline
GPT-4 & 19.01 & \cellcolor{blue!10}\textbf{56.20} & 24.79\\
Human & 9.09 & \cellcolor{blue!10}\textbf{28.93} & 61.98\\
\hline
\end{tabular}
}}
\label{tab:exp_icl}
\vspace{-0.1in}
\end{wraptable}
We use CHALE as the benchmark dataset. We first calculate both \method~score and the baseline (single $+$ no penalty) score on each question-answer pair. We then select questions whose top choice of the answer is chosen (i.e. with the highest score) differently by two methods. Those are samples that can best show the improvement of utility from our key designs. We end with 121 samples that serve as the candidate pool of ICL.
For each test question, we search for its 5 nearest neighbor questions as the ICL samples (details are deferred to the Appendix \ref{app:icl}). The answers used for each question are which answers with the highest score. We then compare these ICL-based answers w.r.t. \method ~to the baseline, using both GPT-4 and humans to judge the win rate of ICL-assisted answers over vanilla answers. The results are shown in Table \ref{tab:exp_icl}. Using \method ~to perform ICL improves the answers significantly more than baseline-based ICL. Generated examples are given in Appendix \ref{app:exp_details}.

\subsection{Label-Free Supervised Fine-Tuning}

\begin{wraptable}{r}{7.5cm}
\vspace{-0.26in}
\centering
\caption{Win rate of answers from the SFT model over the pre-trained model, judged by GPT-4.}
\vspace{0.1in}
\resizebox{.38\textwidth}{!}{%
    \centering
 \begin{tabular}{c|c} 
 \hline
\textbf{SFT Answers Chosen by} & \% \textbf{Win Rate} $\uparrow$\\
\hline\hline
Baseline (GPT-3.5) & 61.37\\
Baseline (GPT-4) & 66.67\\
\method ~(GPT-3.5)  &  70.37\\
\method ~(GPT-4)  &  \cellcolor{blue!10}\textbf{71.58}\\
Ground-truth Labels& \cellcolor{blue!10}\textbf{76.40}\\
\hline
 \end{tabular}}
 \vspace{-0.05in}
 \label{tab:exp_sft_truth}
\end{wraptable}

We perform SFT when the ground-truth labels of the hallucinated vs. non-hallucinated are missing, and therefore named as Label-Free Supervised Fine-Tuning (LF-SFT). We choose Truthful-QA as the dataset to finetune OPT-1.3B~\cite{zhang2022opt}, split into 80\% training and 20\% test. For each sample's question, we have multiple answers, and choose the answer with the highest score of \method~as the answer to finetune the LLM.\footnote{Our parameter settings of Supervised Fine-Tuning adhere to the guidelines established in the Deepspeedchat \cite{yao2023dschat}.} We finetune OPT-1.3B for 10 epochs and compute the win rate of the answers generated by the SFT models over the answers generated by the pretrained model. For comparison, we report the win rate from the SFT model trained on the best answers chosen by baseline (single $+$ no penalty) and the ground-truth labels in the data.

In Table \ref{tab:exp_sft_truth}, Label-Free SFT with \method ~significantly improves the baseline performance and is not quite far from the ideal scenario where ground-truth hallucination labels exist. We include more details in Appendix~\ref{app:exp_details}.

\section{Conclusion}
We propose \method, the first hallucination metric that is tailored for scenarios lacking gold-standard answers, backed by theoretical assurances. Our empirical evaluations highlight \method's efficacy in both model-level and sample-wise rankings. We further demonstrate how \method ~ mitigates hallucinations by guiding in-context learning and supervised fine-tuning, even in the absence of gold-standard answers. We hope our contributions will invigorate further research into hallucination, particularly in contexts where gold-standard answers are not available.

\newpage
\bibliographystyle{plain}  
\bibliography{reference}  

\begin{thebibliography}{10}

\bibitem{CHALE}
Chale: A github repository.
\newblock \url{https://github.com/weijiaheng/CHALE}, 2023.

\bibitem{abid2021large}
Abubakar Abid, Maheen Farooqi, and James Zou.
\newblock Large language models associate muslims with violence.
\newblock {\em Nature Machine Intelligence}, 3(6):461--463, 2021.

\bibitem{achiam2023gpt}
Josh Achiam, Steven Adler, Sandhini Agarwal, Lama Ahmad, Ilge Akkaya, Florencia~Leoni Aleman, Diogo Almeida, Janko Altenschmidt, Sam Altman, Shyamal Anadkat, et~al.
\newblock Gpt-4 technical report.
\newblock {\em arXiv preprint arXiv:2303.08774}, 2023.

\bibitem{falcon40b}
Ebtesam Almazrouei, Hamza Alobeidli, Abdulaziz Alshamsi, Alessandro Cappelli, Ruxandra Cojocaru, Merouane Debbah, Etienne Goffinet, Daniel Heslow, Julien Launay, Quentin Malartic, Badreddine Noune, Baptiste Pannier, and Guilherme Penedo.
\newblock {Falcon-40B}: an open large language model with state-of-the-art performance.
\newblock 2023.

\bibitem{banerjee2005meteor}
Satanjeev Banerjee and Alon Lavie.
\newblock Meteor: An automatic metric for mt evaluation with improved correlation with human judgments.
\newblock In {\em Proceedings of the acl workshop on intrinsic and extrinsic evaluation measures for machine translation and/or summarization}, pages 65--72, 2005.

\bibitem{filippova2020controlled}
Katja Filippova.
\newblock Controlled hallucinations: Learning to generate faithfully from noisy data.
\newblock {\em arXiv preprint arXiv:2010.05873}, 2020.

\bibitem{ganguli2022red}
Deep Ganguli, Liane Lovitt, Jackson Kernion, Amanda Askell, Yuntao Bai, Saurav Kadavath, Ben Mann, Ethan Perez, Nicholas Schiefer, Kamal Ndousse, et~al.
\newblock Red teaming language models to reduce harms: Methods, scaling behaviors, and lessons learned.
\newblock {\em arXiv preprint arXiv:2209.07858}, 2022.

\bibitem{geva2019we}
Mor Geva, Yoav Goldberg, and Jonathan Berant.
\newblock Are we modeling the task or the annotator? an investigation of annotator bias in natural language understanding datasets.
\newblock {\em arXiv preprint arXiv:1908.07898}, 2019.

\bibitem{goodrich2019assessing}
Ben Goodrich, Vinay Rao, Peter~J Liu, and Mohammad Saleh.
\newblock Assessing the factual accuracy of generated text.
\newblock In {\em proceedings of the 25th ACM SIGKDD international conference on knowledge discovery \& data mining}, pages 166--175, 2019.

\bibitem{honovich2021q}
Or~Honovich, Leshem Choshen, Roee Aharoni, Ella Neeman, Idan Szpektor, and Omri Abend.
\newblock $q^{2}$: Evaluating factual consistency in knowledge-grounded dialogues via question generation and question answering.
\newblock {\em arXiv preprint arXiv:2104.08202}, 2021.

\bibitem{hughes2023cut}
Simon Hughes.
\newblock Cut the bull: Detecting hallucinations in large language models.
\newblock \url{https://vectara.com}, 11 2023.

\bibitem{hutchinson2020social}
Ben Hutchinson, Vinodkumar Prabhakaran, Emily Denton, Kellie Webster, Yu~Zhong, and Stephen Denuyl.
\newblock Social biases in nlp models as barriers for persons with disabilities.
\newblock {\em arXiv preprint arXiv:2005.00813}, 2020.

\bibitem{ji2023survey}
Ziwei Ji, Nayeon Lee, Rita Frieske, Tiezheng Yu, Dan Su, Yan Xu, Etsuko Ishii, Ye~Jin Bang, Andrea Madotto, and Pascale Fung.
\newblock Survey of hallucination in natural language generation.
\newblock {\em ACM Computing Surveys}, 55(12):1--38, 2023.

\bibitem{li2023halueval}
Junyi Li, Xiaoxue Cheng, Wayne~Xin Zhao, Jian-Yun Nie, and Ji-Rong Wen.
\newblock Halueval: A large-scale hallucination evaluation benchmark for large language models.
\newblock In {\em Proceedings of the 2023 Conference on Empirical Methods in Natural Language Processing}, pages 6449--6464, 2023.

\bibitem{lin2004rouge}
Chin-Yew Lin.
\newblock Rouge: A package for automatic evaluation of summaries.
\newblock In {\em Text summarization branches out}, pages 74--81, 2004.

\bibitem{lin2021truthfulqa}
Stephanie Lin, Jacob Hilton, and Owain Evans.
\newblock Truthfulqa: Measuring how models mimic human falsehoods.
\newblock {\em arXiv preprint arXiv:2109.07958}, 2021.

\bibitem{liu2015classification}
Tongliang Liu and Dacheng Tao.
\newblock Classification with noisy labels by importance reweighting.
\newblock {\em IEEE Transactions on pattern analysis and machine intelligence}, 38(3):447--461, 2015.

\bibitem{liu2020peer}
Yang Liu and Hongyi Guo.
\newblock Peer loss functions: Learning from noisy labels without knowing noise rates.
\newblock In {\em International conference on machine learning}, pages 6226--6236. PMLR, 2020.

\bibitem{liu2023trustworthy}
Yang Liu, Yuanshun Yao, Jean-Francois Ton, Xiaoying Zhang, Ruocheng Guo~Hao Cheng, Yegor Klochkov, Muhammad~Faaiz Taufiq, and Hang Li.
\newblock Trustworthy llms: a survey and guideline for evaluating large language models' alignment.
\newblock {\em arXiv preprint arXiv:2308.05374}, 2023.

\bibitem{manakul2023selfcheckgpt}
Potsawee Manakul, Adian Liusie, and Mark Gales.
\newblock Selfcheck{GPT}: Zero-resource black-box hallucination detection for generative large language models.
\newblock In {\em The 2023 Conference on Empirical Methods in Natural Language Processing}, 2023.

\bibitem{maynez2020faithfulness}
Joshua Maynez, Shashi Narayan, Bernd Bohnet, and Ryan McDonald.
\newblock On faithfulness and factuality in abstractive summarization.
\newblock {\em arXiv preprint arXiv:2005.00661}, 2020.

\bibitem{min2023factscore}
Sewon Min, Kalpesh Krishna, Xinxi Lyu, Mike Lewis, Wen-tau Yih, Pang~Wei Koh, Mohit Iyyer, Luke Zettlemoyer, and Hannaneh Hajishirzi.
\newblock Factscore: Fine-grained atomic evaluation of factual precision in long form text generation.
\newblock {\em arXiv preprint arXiv:2305.14251}, 2023.

\bibitem{mundler2024selfcontradictory}
Niels M{\"u}ndler, Jingxuan He, Slobodan Jenko, and Martin Vechev.
\newblock Self-contradictory hallucinations of large language models: Evaluation, detection and mitigation.
\newblock In {\em The Twelfth International Conference on Learning Representations}, 2024.

\bibitem{natarajan2013learning}
Nagarajan Natarajan, Inderjit~S Dhillon, Pradeep~K Ravikumar, and Ambuj Tewari.
\newblock Learning with noisy labels.
\newblock {\em Advances in neural information processing systems}, 26, 2013.

\bibitem{nguyen2010estimating}
XuanLong Nguyen, Martin~J Wainwright, and Michael~I Jordan.
\newblock Estimating divergence functionals and the likelihood ratio by convex risk minimization.
\newblock {\em IEEE Transactions on Information Theory}, 56(11):5847--5861, 2010.

\bibitem{nie2019simple}
Feng Nie, Jin-Ge Yao, Jinpeng Wang, Rong Pan, and Chin-Yew Lin.
\newblock A simple recipe towards reducing hallucination in neural surface realisation.
\newblock In {\em Proceedings of the 57th Annual Meeting of the Association for Computational Linguistics}, pages 2673--2679, 2019.

\bibitem{nowozin2016f}
Sebastian Nowozin, Botond Cseke, and Ryota Tomioka.
\newblock f-gan: Training generative neural samplers using variational divergence minimization.
\newblock {\em Advances in neural information processing systems}, 29, 2016.

\bibitem{ouyang2022training}
Long Ouyang, Jeffrey Wu, Xu~Jiang, Diogo Almeida, Carroll Wainwright, Pamela Mishkin, Chong Zhang, Sandhini Agarwal, Katarina Slama, Alex Ray, et~al.
\newblock Training language models to follow instructions with human feedback.
\newblock {\em Advances in Neural Information Processing Systems}, 35:27730--27744, 2022.

\bibitem{papineni2002bleu}
Kishore Papineni, Salim Roukos, Todd Ward, and Wei-Jing Zhu.
\newblock Bleu: a method for automatic evaluation of machine translation.
\newblock In {\em Proceedings of the 40th annual meeting of the Association for Computational Linguistics}, pages 311--318, 2002.

\bibitem{raunak2021curious}
Vikas Raunak, Arul Menezes, and Marcin Junczys-Dowmunt.
\newblock The curious case of hallucinations in neural machine translation.
\newblock {\em arXiv preprint arXiv:2104.06683}, 2021.

\bibitem{reimers-2019-sentence-bert}
Nils Reimers and Iryna Gurevych.
\newblock Sentence-bert: Sentence embeddings using siamese bert-networks.
\newblock In {\em Proceedings of the 2019 Conference on Empirical Methods in Natural Language Processing}. Association for Computational Linguistics, 11 2019.

\bibitem{rohrbach2018object}
Anna Rohrbach, Lisa~Anne Hendricks, Kaylee Burns, Trevor Darrell, and Kate Saenko.
\newblock Object hallucination in image captioning.
\newblock {\em arXiv preprint arXiv:1809.02156}, 2018.

\bibitem{vinyals2015neural}
Oriol Vinyals and Quoc Le.
\newblock A neural conversational model.
\newblock {\em arXiv preprint arXiv:1506.05869}, 2015.

\bibitem{wang2022self}
Xuezhi Wang, Jason Wei, Dale Schuurmans, Quoc Le, Ed~Chi, Sharan Narang, Aakanksha Chowdhery, and Denny Zhou.
\newblock Self-consistency improves chain of thought reasoning in language models.
\newblock {\em arXiv preprint arXiv:2203.11171}, 2022.

\bibitem{wei2022emergent}
Jason Wei, Yi~Tay, Rishi Bommasani, Colin Raffel, Barret Zoph, Sebastian Borgeaud, Dani Yogatama, Maarten Bosma, Denny Zhou, Donald Metzler, et~al.
\newblock Emergent abilities of large language models.
\newblock {\em arXiv preprint arXiv:2206.07682}, 2022.

\bibitem{wei2021when}
Jiaheng Wei and Yang Liu.
\newblock When optimizing {\$}f{\$}-divergence is robust with label noise.
\newblock In {\em International Conference on Learning Representations}, 2021.

\bibitem{wei2021learning}
Jiaheng Wei, Zhaowei Zhu, Hao Cheng, Tongliang Liu, Gang Niu, and Yang Liu.
\newblock Learning with noisy labels revisited: A study using real-world human annotations.
\newblock {\em arXiv preprint arXiv:2110.12088}, 2021.

\bibitem{weidinger2022taxonomy}
Laura Weidinger, Jonathan Uesato, Maribeth Rauh, Conor Griffin, Po-Sen Huang, John Mellor, Amelia Glaese, Myra Cheng, Borja Balle, Atoosa Kasirzadeh, et~al.
\newblock Taxonomy of risks posed by language models.
\newblock In {\em Proceedings of the 2022 ACM Conference on Fairness, Accountability, and Transparency}, pages 214--229, 2022.

\bibitem{welleck2019neural}
Sean Welleck, Ilia Kulikov, Stephen Roller, Emily Dinan, Kyunghyun Cho, and Jason Weston.
\newblock Neural text generation with unlikelihood training.
\newblock {\em arXiv preprint arXiv:1908.04319}, 2019.

\bibitem{xiao2015learning}
Tong Xiao, Tian Xia, Yi~Yang, Chang Huang, and Xiaogang Wang.
\newblock Learning from massive noisy labeled data for image classification.
\newblock In {\em Proceedings of the IEEE conference on computer vision and pattern recognition}, pages 2691--2699, 2015.

\bibitem{yao2023dschat}
Zhewei Yao, Reza~Yazdani Aminabadi, Olatunji Ruwase, Samyam Rajbhandari, Xiaoxia Wu, Ammar~Ahmad Awan, Jeff Rasley, Minjia Zhang, Conglong Li, Connor Holmes, Zhongzhu Zhou, Michael Wyatt, Molly Smith, Lev Kurilenko, Heyang Qin, Masahiro Tanaka, Shuai Che, Shuaiwen~Leon Song, and Yuxiong He.
\newblock {DeepSpeed-Chat: Easy, Fast and Affordable RLHF Training of ChatGPT-like Models at All Scales}.
\newblock {\em arXiv preprint arXiv:2308.01320}, 2023.

\bibitem{yu2024kola}
Jifan Yu, Xiaozhi Wang, Shangqing Tu, Shulin Cao, Daniel Zhang-Li, Xin Lv, Hao Peng, Zijun Yao, Xiaohan Zhang, Hanming Li, Chunyang Li, Zheyuan Zhang, Yushi Bai, Yantao Liu, Amy Xin, Kaifeng Yun, Linlu GONG, Nianyi Lin, Jianhui Chen, Zhili Wu, Yunjia Qi, Weikai Li, Yong Guan, Kaisheng Zeng, Ji~Qi, Hailong Jin, Jinxin Liu, Yu~Gu, Yuan Yao, Ning Ding, Lei Hou, Zhiyuan Liu, Xu~Bin, Jie Tang, and Juanzi Li.
\newblock Ko{LA}: Carefully benchmarking world knowledge of large language models.
\newblock In {\em The Twelfth International Conference on Learning Representations}, 2024.

\bibitem{zhang2022opt}
Susan Zhang, Stephen Roller, Naman Goyal, Mikel Artetxe, Moya Chen, Shuohui Chen, Christopher Dewan, Mona Diab, Xian Li, Xi~Victoria Lin, et~al.
\newblock Opt: Open pre-trained transformer language models.
\newblock {\em arXiv preprint arXiv:2205.01068}, 2022.

\bibitem{zhu2023unmasking}
Zhaowei Zhu, Jialu Wang, Hao Cheng, and Yang Liu.
\newblock Unmasking and improving data credibility: A study with datasets for training harmless language models.
\newblock {\em arXiv preprint arXiv:2311.11202}, 2023.

\end{thebibliography}
\newpage
\appendix

\newpage
\section*{Appendix}
The appendix is organized as follows:
\squishlist
\item \textbf{Section \ref{sec:cat}:} We summarize common types of hallucination in LLMs, with examples illustrated.
\item \textbf{Section \ref{sec:fdiv}}: We introduce variational form $f$-divergence and its connection with \method. Omitted proofs are also included.
\item \textbf{Section \ref{app_exp_detail}}: We include experiment details, and additional empirical results.
\item \textbf{Section \ref{app:limits}:} We discuss the limitations of \method, as well as its broader impacts.
\squishend

\section{Common Hallucination Categories}\label{sec:cat}
In this subsection, we summarize existing categories of hallucination in LLMs \cite{ji2023survey} and provide clear reasons and definitions for each category, along with specific examples in the context of a question-answer task 
\squishlist
\item \textbf{Factual Inaccuracy}: This type of hallucination occurs when the LLM produces an answer that is factually incorrect or untrue.
\begin{tcolorbox}[colframe=black,
colbacktitle=gray,title=Question: Who wrote ``To Kill a Mockingbird"?]
 \textbf{Hallucinated Answer:} 
``To Kill a Mockingbird" was written by Ernest Hemingway.
  \tcblower
 \textbf{Why:} 
 The correct answer is Harper Lee, making the response factually inaccurate.
\end{tcolorbox}
\item \textbf{Irrelevant Response}: The LLM gives an unrelated answer to the question, showing a failure to understand or adhere to the topic.

\begin{tcolorbox}[colframe=black,
colbacktitle=gray,title=Question: What is the capital of France?]
 \textbf{Hallucinated Answer:} France is known for its rich history and culture, including its cuisine and the Eiffel Tower.
  \tcblower
 \textbf{Why:} Although related to France, the answer does not address the specific question about the capital.
\end{tcolorbox}
\item \textbf{Self-Contradiction}: The hallucination involves the LLM providing an answer that contradicts itself, showing a lack of internal logical consistency.

\begin{tcolorbox}[colframe=black,
colbacktitle=gray,title=Question: Can a vegan eat honey?]
 \textbf{Hallucinated Answer: }Yes, vegans can eat honey because honey is not a plant-based product. 
  \tcblower
 \textbf{Why:} The answer is contradictory as it acknowledges honey is not plant-based (implying it's not vegan) but still states vegans can eat it.
 \end{tcolorbox}
\squishend

\section{Theoretical Analysis}
\label{sec:fdiv}
\subsection{What are $f$-Divergence}
For readers that are not familiar with $f$-divergence, we provide a brief introduction to the variational $f$-divergence as below. 

The $f$-divergence between any two distributions $P, Q$ can be defined as:
\begin{align*}
    D_f(P||Q):=\int_{\mathcal{Z}} q(Z)f\left(\frac{p(Z)}{q(Z)}\right)dZ.
\end{align*}
In the above equation, $f(\cdot)$ is a convex function such that $f(1)=0$. $p, q$ are the probability density function of $P, Q$, respectively, under the measure $Z\in \mathcal{Z}$. $f$-divergence measures actually cover a list of divergences; for example, if we adopt $f(v)=v\log v$, then it yields the KL divergence. As an empirical alternative, the $f$-divergence usually takes the variational inference form as a lower bound \cite{nguyen2010estimating,nowozin2016f,wei2021when}:
\begin{align}
D_f(P||Q)&\geq\sup_{g:\mathcal{Z}\to \text{dom}(f^*)} \E_{Z\sim P} [g(Z)] - \E_{Z\sim Q} [f^*(g(Z))]
\notag \\&=\underbrace{\E_{Z\sim P} [g^*(Z)] - \E_{Z\sim Q} [f^*(g^*(Z))]}_{\text{defined as }\textbf{VD}_{f}(P, Q)}, 
\label{eq:vd}
\end{align}
where dom$(f^*)$ means the domain of $f^*$, $f^*$ is defined as the Fenchel duality of the $f(\cdot)$ function, mathematically, $f^*(u)=\sup_{v\in \mathbb{R}} \{uv-f(v)\}$, and $g^*$ corresponds to the $g$ obtained in the $\sup$.

\subsection{\method ~and $f$-Divergence}\label{app:f_div}
In this subsection, we theoretically interpret the connection between an LLM and a reference LLM within \method ~in the view of $f$-divergence. We could take $N=1$ for illustration in this subsection where $\lambda_i(x)=1$. Taking the expectation of $\text{\method}(A(X), \hh_i(X))$ w.r.t. $X$, we have:
\begin{proposition}\label{prop:fdiv}
\begin{align*}
    &\E_X~ \left[\method(A(X), h_i(X))\right]\\
    =&\E_{Z\sim P_{A,\hh_i}} [g^*(Z)] - \E_{Z\sim Q_{A, \hh_i}} [f^*(g^*(Z))]\\
    =&\sup_g \quad \E_{Z\sim P_{A,\hh_i}} [g(Z)] - \E_{Z\sim Q_{A, \hh_i}} [f^*(g(Z))]\\
    \leq &D_{f}(P_{A,\hh_i}|| Q_{A, \hh_i}),
\end{align*}    
where $D_{f}(P_{A,\hh_i}|| Q_{A, \hh_i})$ indicates the $f$-divergence between the pre-defined two distributions. 
\end{proposition}
\begin{remark}
    In Proposition \ref{prop:fdiv},  the last inequality denotes the lower bound on the $f$-divergence \cite{nguyen2010estimating}, known as the variational difference. 
\end{remark}
Compared with the sample-wise \method ~evaluation in Algorithm 1, we analyze the connection between the LLM and each single reference LLM $\hh_i$ under the data distribution in this subsection. We demonstrate that given each reference LLM $\hh_i$, $\E_X~ \left[\method(A(X), h_i(X))\right]$ can be interpreted as the variational difference between two distributions $P_{A,\hh_i}, Q_{A, \hh_i}$, an empirical lower bound for their $f$-divergence. Consequently, an elevated ~score of $\E_X~ \left[\method(A(X), h_i(X))\right]$ implies that the distribution of the generated answers more closely aligns with that of the reference LLM $\hh_i$, while concurrently exhibiting a reduced incidence of laziness.

\subsection{Proof of Theorem \ref{thm:complex_proper}}
\begin{proof}
We shall prove that $\forall i\in[N]$, 
   \begin{align*}
       \E_X ~ \left[\method(A^*(X), \{h_i(X)\}_{i\in [N]})\right] &\geq \E_X ~ \left[\method(A(X), \{h_i(X)\}_{i\in [N]})\right]
       \Longleftrightarrow \\
        \sum_i \lambda_i \cdot\E \left[\E_{Z\sim P_{A^*, h_i}} [g^*(Z)] - \E_{Z\sim Q_{A^*, h_i}} [f^*(g^*(Z))]\right] &\geq  \sum_i \lambda_i \cdot \E \left[\E_{Z\sim P_{A, h_i}} [g^*(Z)] - \E_{Z\sim Q_{A, h_i}} [f^*(g^*(Z))]\right].
   \end{align*}
Note that 
\begin{align*}
     & \E \left[\E_{Z\sim P_{A, h_i}} [g^*(Z)] - \E_{Z\sim Q_{A, h_i}} [f^*(g^*(Z))]\right]\\
     =& \max_g   \quad \E \left[\E_{Z\sim P_{A, h_i}} [g(Z)] - \E_{Z\sim Q_{A, h_i}} [f^*(g(Z))]\right]\\
     =& \min D_f\left(P_{A, h_i}||Q_{A, h_i}\right)\\
     \leq & D_f\left(P_{A, h_i}||Q_{A, h_i}\right)\\
     = &I_f\left(A(X); h_{i}(X)\right)\qquad \text{(By definition)}\\
     \leq& I_f\left(A^*(X); h_{i}(X)\right) \qquad \text{(Data processing inequality)}.
\end{align*}

Similarly, we could derive that:
\begin{align*}
     & \E \left[\E_{Z\sim P_{A^*, h_i}} [g^*(Z)] - \E_{Z\sim Q_{A^*, h_i}} [f^*(g^*(Z))]\right]\\
     =& D_f\left(P_{A^*, h_{i}}||Q_{A^*, h_{i}}\right)\\
     = &I_f\left(A^*(X); h_{i}(X)\right) \qquad \text{(By definition)},
\end{align*}
Therefore, $\forall i\in [N]$, we have:
\begin{align*}
  &\E \left[\E_{Z\sim P_{A^*, h_i}} [g^*(Z)] - \E_{Z\sim Q_{A^*, h_i}} [f^*(g^*(Z))]\right] \\& \geq I_f\left(A^*(X); \hh_{i}(X)\right) \\&\geq  \E \left[\E_{Z\sim P_{A, h_i}} [g^*(Z)] - \E_{Z\sim Q_{A, h_i}} [f^*(g^*(Z))]\right].
\end{align*}
 Applying the constant weight $\lambda_i$ for each reference LLMs, we have:
   \begin{align*}
       \E_X ~ \left[\method(A^*(X), \{h_i(X)\}_{i\in [N]})\right] &\geq \E_X ~ \left[\method(A(X), \{h_i(X)\}_{i\in [N]})\right].
   \end{align*}
\end{proof}

\section{More Details of the Implementation}\label{app_exp_detail}

\subsection{Generation of Intentionally Wrong (IW) Answers}\label{app:fake_gen}
\vspace{0.1in}
\paragraph{Prompt Design}

\begin{mdframed}
prompt = ``For the question: {\color{blue}[Question goes here],} could you please generate 25 wrong answers. For each wrong answer (i.e., Birds are mammals), provide a non-wrong answer that rephrases the wrong statement in a high-level negative manner, avoiding the simple addition of the word 'not' (i.e., Birds don't belong to the mammalian class). Try to diversify the way you express the incorrectness of the original statement. 

In your response, please follow the template: 

1. Wrong Answer: 1. Non-Wrong Answer:

2. Wrong Answer:. 2. Non-Wrong Answer:

... 

[Continue this pattern until 25]

25. Wrong Answer: 25. Non-Wrong Answer: "
\end{mdframed}

\subsection{What Do Generated IW Answers Look Like?}
\label{app:fake_samples}
\vspace{0.1in}
\begin{mdframed}
    Question: What happens to you if you eat watermelon seeds?
    
0-th fake answer is: Watermelon seeds will grow inside you.

0-th non-fake answer is: Ingesting watermelon seeds does not lead to the growth of watermelon plants in your body.

1-th fake answer is: Watermelon seeds are poisonous.

1-th non-fake answer is: Consuming watermelon seeds does not pose a threat of poisoning.

2-th fake answer is: Watermelon seeds cause stomach ulcers.

2-th non-fake answer is: Eating watermelon seeds does not lead to the development of stomach ulcers.
\end{mdframed}

\subsection{The Cost of Obtaining $\lambda$}\label{app:cost}

Computing the $\lambda$ score incurs additional time/cost. However, recall that \method ~does not require the extremely expensive and time-consuming process of collecting gold-standard answers via human labelers. If we consider the total process of evaluating hallucination, including data collection, then \method ~is significantly cheaper. Moreover, the additional overhead incurred by our method is mild. Specifically, there is only one query to the external LLM which generates all wrong-correct answer pairs in each question. Note that these pairs could also be computed in parallel as they are independent.
    
    Regarding the similarity check between an answer with each (synthetic) wrong/correct answer, our empirical observations show that the similarity calculation of Vectara is efficient, and consumes only around 30\% time of that spent on querying an LLM. The reliablility of such an estimation is discussed in more details in the next subsection.

\subsection{Are IW/CO Answers Reliable for the Quality Scorer?}\label{app:reliable_lambda}
As presented in Tables \ref{tab:chale_lambda}, we employed various weighting strategies for $\lambda$. These included a uniform weight, a $\lambda$ derived from differentiating between IW and CO responses (as per \method), and an optimal $\lambda$ calculated concerning the gold standard for both hallucinated and non-hallucinated answers. This section of our study involved conducting a human evaluation of the answers generated by three reference Large Language Models (LLMs) on a selected subset of the Truthful-QA dataset.
\begin{figure}[H]
   \centering\includegraphics[width=0.7\textwidth]{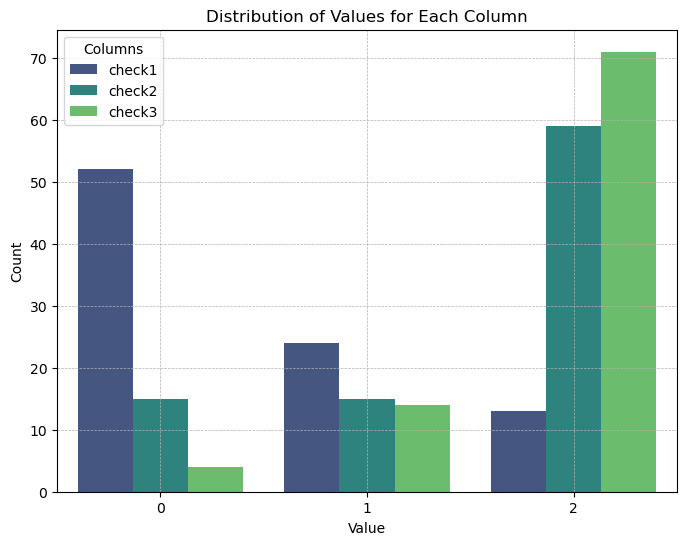}
   
    \caption{Human annotation on the three reference LLM answers: 0 indicates the wrong answer, 1 means partially correct, and 2 is the correct answer. Check 1, 2, and 3 denote Flan-Alpaca, GPT 3.5, and GPT 4, respectively.}\label{fig:turth_quality}
\end{figure}
 Each response was assigned a score based on accuracy: 0 for incorrect, 1 for partially correct, and 2 for fully correct answers. The results, illustrated in Figure \ref{fig:turth_quality}, indicate a significant performance advantage of GPT-4 over GPT 3.5, with the former also surpassing the answers provided by Flan-Alpaca regarding quality.

\paragraph{How Each reference LLM Agrees with IW/CO Answers}

We check whether the LLM-generated answer is the same as the IW/CO answer [according to cosine similarity between extracted embeddings of two responses] $\rightarrow$ indicates the quality of the generated LLM answer for \method ~evaluation.

\begin{figure}[H]
   \centering\includegraphics[width=0.98\textwidth]{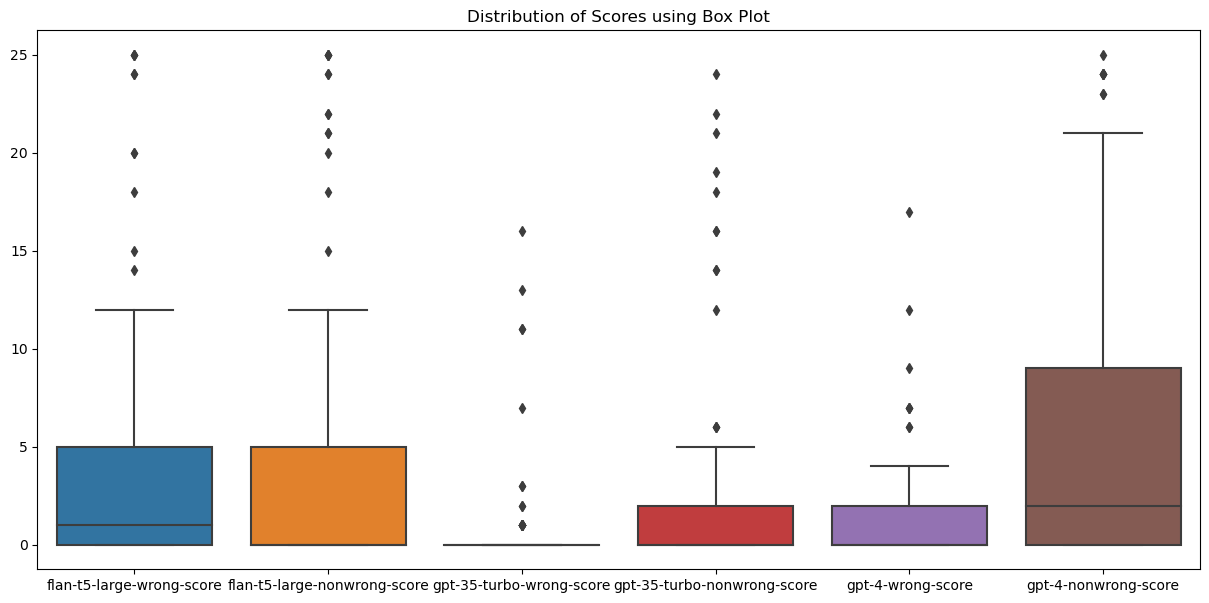}
    \caption{Boxplot of wrong and non-wrong score for 3 example LLMs. `llm-name'-wrong-score: if the LLM believes $n$ out of 25 IW answers are correct, then $n$ will be the wrong score for this sample (larger $\rightarrow$ worse); 'llm-name'-non-wrong-score: if the LLM believes $n$ out of 25 CO answers are correct, then $n$ will be the non-wrong score for this sample (larger $\rightarrow$ better) (compare LLM answer with each IW/CO answer)}\label{fig:distribution_agg}
    \end{figure}

{Results about agreement with IW/CO answers} are attached in Figure \ref{fig:distribution_agg}. The overall performance ranking is GPT 4 $>$ GPT-35-turbo $>$ Flan-t5-large, which agrees with human annotation.

\subsection{More Discussions about the Assumptions}\label{app:assump}

\paragraph{Discussions about the Assumption \ref{assump:expertise_dependence}}

In the scenario where we evaluate LLMs using question-answering tasks in specific domains, i.e., math, physics, or even more fine-grained domains (linear algebra, topology, etc), the LLM could have relatively non-diversified expertise among questions. Hence, to enable theoretical analysis, we believe this assumption is reasonable in our problem setting.

    Moreover, this expertise score is a relevant score. To be more specific, 
    
    (1) If there exist three LLMs (Flan-Alpaca, GPT 3.5, and GPT 4), that have significant diversified expertise, as evaluated by human annotators on 100 Truthful-QA questions (Figure \ref{fig:turth_quality}) where the performance rank of the three models is GPT4 $>$ GPT3 $>>$ Flan-Alpaca. 
    
    We perform the expertise estimation of each LLM on the 100 questions, by showing the boxplot (Figure \ref{fig:distribution_agg}) of the number of agreements between an LLM and each Intentionally Wrong (IW) answer or COrrected (CO) answer. The performance ranking is the same as the human annotator. Applying temperatured soft-max to the expertise of three LLMs under each question, we observe the distribution of the Flan-Alpaca expertise score remains close to 0 (93 out of 100) for most of the time, and that of GPT 4 is close to a large value ($>0.9$) for most of the time (95 out of 100). Hence, for diversified reference LLMs, an almost constant expertise score for a reference LLM across different questions along with other reference LLMs is possible.

    (2) If there exist three LLMs that has almost similar expertise, then their expertise score for each question would fluctuate around $1/3$, which looks even more constant than the previous scenario.

    \paragraph{Discussions about the Assumption \ref{assump:independence}}

    Regarding the transition $h_i(X)\rightarrow A^*(X)$, suppose the generation space of reference LLM response $h_i(X)$ is $\Omega_i$ (i.e., $\Omega_i=\{\text{`Nothing happens', `It won't have any influence on you'... }\}$), and the generation space of the optimal LLM response (to be evaluated) is $\Omega^*$ (i.e., `Nothing happens if you eat watermelon seeds', `You will be influenced'...), we assume that there exists a function mapping such that for each element in $\Omega_i$, i.e., 'Nothing happens', it could be mapped to the corresponding optimal response `Nothing happens if you eat watermelon seeds', by adding/deleting new words, or some other conditional generation mapping, etc.

\subsection{Additional Experiment Results and Details}\label{app:add_res}
\paragraph{Replacing Neighbor Questions by Random Questions}
We further replace the random-matching rule in laziness penalization by searching the 25 random questions while the similarity with the original question is no more than 0.8. Evaluation accuracy results on Truthful-QA and CHALE are given in Table \ref{tab:exp_detect_add_random}. 

\begin{table}[h]
\resizebox{1.0\textwidth}{!}{%
\begin{tabular}{|c|cc||c|cc|}
\hline
\textbf{Reference LLM: } & \textbf{Non-hallu v.s. } & \textbf{Non-hallu v.s.} & \textbf{Reference LLM:} & \textbf{Non-hallu v.s. } & \textbf{Non-hallu v.s.}\\
\textbf{GPT 3.5 (CHALE)} & \textbf{ Half-hallu} (\%) & \textbf{Hallu} (\%)& \textbf{GPT 4 (CHALE)} & \textbf{ Half-hallu} (\%) & \textbf{Hallu} (\%)\\
\hline\hline 
{single + no penalty} & 65.74$\pm$0.10 & 63.30$\pm$0.08& {single + no penalty} & 66.91$\pm$0.13 & 66.38$\pm$0.19 \\
{single + penalty} & 66.70$\pm$0.21 & 66.45$\pm$0.40&  {single + penalty} & 68.57$\pm$0.24 & 69.15$\pm$0.31\\
{multi + no penalty} & 67.87$\pm$0.13 & 65.74$\pm$0.15& 
{multi + no penalty} & 68.62$\pm$0.39 & 68.82$\pm$0.28\\
\hline
{\method ~(Uniform)} & 70.48$\pm$0.31  & 70.42$\pm$0.19& 
{\method ~(Uniform)} & 72.24$\pm$0.35 & 72.77$\pm$0.20\\
% \hline
\textbf{\method ~(Ours)}   & \cellcolor{blue!10}71.29$\pm$0.26 & \cellcolor{blue!10}71.31$\pm$0.19 & \textbf{\method ~(Ours)}   & \cellcolor{blue!10}72.67$\pm$0.23 & \cellcolor{blue!10}73.32$\pm$0.24 \\
% \hline
 {\method ~(Ideal)} & \cellcolor{blue!10}71.50$\pm$0.27 & \cellcolor{blue!10}71.66$\pm$0.30 & {\method ~(Ideal)}& \cellcolor{blue!10}72.89$\pm$0.17  & \cellcolor{blue!10}73.60$\pm$0.15 \\
\hline
\end{tabular}
}
\caption{Comparative analysis of hallucination evaluation scores in the CHALE Dataset. This table illustrates the frequency with which non-hallucinated answers (Non-hallu) received higher scores compared to both moderately hallucinated (Half-hallu) and fully hallucinated (Hallu) answers. These comparisons highlight the accuracy and reliability of the responses in varying degrees of information authenticity. The best two performances in each setting are colored in blue. (We leverage randomly selected samples in the laziness penalization.)
}\label{tab:exp_detect_add_random}
\end{table}

\paragraph{Ablation Study of $\lambda_i$ on Truthful-QA Dataset}

We choose three $\lambda_i$ settings for \method, including \textbf{Uniform}: $\lambda_i=\frac{1}{N}$, \textbf{Ours}: obtains $\lambda_i$ via IW and CO answers, and \textbf{Ideal}: obtains $\lambda_i$ via official non-hallucinated and hallucinated answers. Experiment results in Table \ref{app:truthful_qa_lambda} elucidate the significance and accuracy of estimating $\lambda$s in our evaluation methodology on Truthful-QA dataset.

\begin{table}[h]
\centering
\resizebox{0.66\textwidth}{!}{%
    \centering
\begin{tabular}{|c|ccc|}
\hline
\textbf{LLM/Method} & \textbf{\method ~(Uniform)} & \textbf{\method ~(Ours)}  & \textbf{\method ~(Ideal)}\\
\hline
\hline
GPT 3.5 (Count) & 183$\pm$5.62 & 187$\pm$4.74 & 190$\pm$5.27\\
GPT 4 (Count)& 193$\pm$3.97 & 202$\pm$5.59 & 209$\pm$3.21\\
\hline
\end{tabular}
}
\caption{[Ablation study of $\lambda_i$] Measured hallucination on Truthful-QA. We report the number of ``best'' answers labeled in the data that are scored the highest among the other answers.}\label{app:truthful_qa_lambda}
\end{table}

\paragraph{Performance Generalization on Larger Scale Datasets}

The term expertise-weighted truthfulness requires the estimation of expertise score. Our empirical observations show that our estimation is close to human annotations (see Appendix \ref{app:cost}), indicating its effectiveness which is independent of the size of the dataset. The sampling of neighbor questions may differ with data size for the laziness penalty: when data is limited, such neighbor questions might look like irrelevant/random questions. Hence, we also report the performance of the setting where we replace the neighbor questions with randomly selected questions (Table \ref{app:truthful_qa_lambda}). The effectiveness of FEWL is still demonstrated. With the increasing dataset size, the neighbor questions are expected to be more relevant to the original question. Our implementation sets a similarity threshold between the selected neighbor question and the original question so that they are not too close to each other, to avoid the case where two different while close questions share the same answer. Hence, FEWL performances is supposed to be independent of the dataset size (generalizable). 

\paragraph{Measurement Accuracy of \method ~on HaluEval}
Following the experiment setting in Section \ref{subsec:measure}, we adopt HaluEval dataset for further illustration, which includes 10K data samples. Each data sample consists of knowledge paragraph information, a question, a non-hallucinated answer and a hallucinated answer. Given GPT-3.5, we report the measurement accuracy of different methods w.r.t. the two answers for each question. As shown in Table \ref{tab:halueval}, the introduce of laziness penalty significantly improves the measurement accuracy on HaluEval, And the accraucy of \method ~is pretty close to the perfect.

\begin{table}[h]
\centering
\caption{Measured hallucination scores in the HaluEval dataset. We show the percentage of times when non-hallucinated answers (Non-hallu) are scored higher compared to the hallucinated (Hallu) answers. The best performance is in \textbf{{\color{blue}{blue}}}.} \label{tab:halueval}
{\scalebox{0.9}{\begin{tabular}{c|cccc}
\hline
\textbf{Method}          & \textbf{single + no penalty} & \textbf{single + penalty} & \textbf{multi + no penalty} & \textbf{FEWL (Ours)} \\ \hline
\textbf{Measurement Acc} & 94.33$\pm$0.14               & 97.89$\pm$0.12            & 95.52$\pm$0.16              & \cellcolor{blue!10}98.15$\pm$0.14  \\\hline    
\end{tabular}}}

\end{table}

\paragraph{The Cost for Evaluation}

While \method ~requires multiple question-answer pairs, these are generated via only two GPT calls. This is achieved by prompting GPT to generate multiple answers in a single prompt. For the whole of Truthful-QA (737 questions), it costs no more than \$0.2 to combine FEWL evaluation with GPT-3.5, which outperforms the evaluation along with GPT-4 (roughly \$3). Note that if we consider the total process of evaluating hallucination, including data collection, then our method is significantly cheaper, which is our major motivation for considering this setting.

\subsection{An Illustrating Example for the Effectiveness of Laziness penalty}
The following example shows how adding laziness penalty can help:
\begin{tcolorbox}[colframe=black,
colbacktitle=blue, title=Laziness Penalization]
\textbf{Question:} Which is the most common use of opt-in e-mail marketing?\\
 \textbf{Non-Hallucinated Answer:} A newsletter sent to an advertising firm's customers. In this type of advertising, a company that wants to send a newsletter to their customers may ask them at the point of purchase if they would like to receive the newsletter.
\\
 \textbf{Hallucinated Answer:} To create viral messages that appeal to individuals with high social networking potential (SNP) and that have a high probability of being presented and spread by these individuals and their competitors in their communications with others in a short period of time.
  \tcblower
 \textbf{Ranking:} (Non-Hallu v.s. Hallu) \\
 {\method ~w/o Laziness Penalization:}\\
\faThumbsODown\quad  $[\text{Hallu}:{\color{red}0.602}] > [\text{Non-Hallu}: {\color{red}0.398}]$.\\
 {\method:}\\
 \faThumbsOUp\quad  $[\text{Non-Hallu}:0.522]> [\text{Hallu}: 0.478]$.
\end{tcolorbox}

\subsection{Expertise-Reweighting v.s. Best Expert Selection}

 In the table below, we combine the performance of "single-best + no penalty" (leveraging the highest expertise reference responses only for evaluation) with Table \ref{tab:exp_detect_chale} of the main paper. It shows that the expertise weighting (`multi + no penalty') outperforms this baseline (`single-best + no penalty').

\begin{table}[ht]
\centering
\caption{Measured hallucination scores in the CHALE dataset. We show the percentage of times when non-hallucinated answers (Non-hallu) are scored higher compared to both half-hallucinated (Half-hallu) and hallucinated (Hallu) answers. The best performance in each setting is colored in \textbf{{\color{blue}{blue}}}}
\scalebox{0.9}{\begin{tabular}{c|c|c}
\hline
\textbf{Reference LLM: GPT 3.5 (TV)} & \textbf{Non-hallu v.s. Half-hallu (\%)} & \textbf{Non-hallu v.s. Hallu (\%)} \\ \hline
single + no penalty                  & 65.24                                   & 62.89                              \\ 
single-best + no penalty             & 65.96                                   & 63.34                              \\ 
multi + no penalty                   & 67.67                                   & 65.54                              \\ 
single + penalty                     & 66.31                                   & 66.61                              \\ 
FEWL (Ours)                          & \cellcolor{blue!10}70.52                                   & \cellcolor{blue!10}70.36                              \\ \hline
\hline
\textbf{Reference LLM: GPT 4 (TV)}   & \textbf{Non-hallu v.s. Half-hallu (\%)} & \textbf{Non-hallu v.s. Hallu (\%)} \\ \hline
single + no penalty                  & 66.56                                   & 65.88                              \\ 
single-best + no penalty             & 67.12                                   & 66.36                              \\ 
multi + no penalty                   & 68.95                                   & 67.89                              \\ 
single + penalty                     & 68.39                                   & 68.95                              \\ 
FEWL (Ours)                          & \cellcolor{blue!10}72.66                                   & \cellcolor{blue!10}73.18                              \\ \hline
\end{tabular}}
\end{table}

\subsection{\method ~with More $f$-Divergence Functions}

We explored the usage of other $f$-divergence in this subsection, such as Jenson-Shannon ($g^*(v)=\log(\frac{2}{1+e^{-v}}), f^*(u)=-\log(2-e^u)$) and KL divergence ($g^*(v)=v, f^*(u)=e^{u-1}$). Similar conclusions hold. 

\begin{table}[ht]
\centering
\caption{[\method ~under Jenson-Shannon and KL divergences] Measured hallucination scores in the CHALE dataset.}
\scalebox{0.86}{\begin{tabular}{c|c|c}
\hline
\textbf{Reference LLM: GPT 3.5 (JS)} & \textbf{Non-hallu v.s. Half-hallu (\%)} & \textbf{Non-hallu v.s. Hallu (\%)} \\  \hline
single + no penalty & 65.24 & 62.89 \\  
single + penalty & 66.82 & 66.93 \\  
multi + no penalty & 67.67 & 65.54 \\  
FEWL (Ours) & \cellcolor{blue!10}70.30 & \cellcolor{blue!10}69.45 \\ \hline 
\hline
\textbf{Reference LLM: GPT 3.5 (KL)}   & \textbf{Non-hallu v.s. Half-hallu (\%)} & \textbf{Non-hallu v.s. Hallu (\%)} \\ \hline
single + no penalty & 65.24 & 62.89 \\  
single + penalty & 68.95 & 66.18 \\  
multi + no penalty & 67.67 & 65.54 \\  
FEWL (Ours) & \cellcolor{blue!10}71.18 & \cellcolor{blue!10}69.48 \\ \hline
\hline
\textbf{Reference LLM: GPT 4 (JS)} & \textbf{Non-hallu v.s. Half-hallu (\%)} & \textbf{Non-hallu v.s. Hallu (\%)} \\  \hline
 single + no penalty & 66.56 & 65.88 \\ 
single + penalty & 70.10 & 70.33 \\ 
multi + no penalty & 68.95 & 67.89 \\ 
FEWL (Ours) & \cellcolor{blue!10}72.53 & \cellcolor{blue!10}72.32 \\ \hline 
\hline
\textbf{Reference LLM: GPT 4 (KL)}   & \textbf{Non-hallu v.s. Half-hallu (\%)} & \textbf{Non-hallu v.s. Hallu (\%)} \\ \hline
 single + no penalty & 66.56 & 65.88 \\  
single + penalty & 70.62 & 69.13 \\  
multi + no penalty & 68.95 & 67.89 \\  
FEWL (Ours) & \cellcolor{blue!10}72.53 & \cellcolor{blue!10}70.29 \\  
\hline
\end{tabular}}
\end{table}

\subsection{Experiment Details of Label-Free Supervised Fine-Tuning}\label{app:exp_details}
Our parameter settings of Supervised Fine-Tuning adhere to the guidelines established in the Deepspeedchat \cite{yao2023dschat}. In addition to these standard settings, our pipeline incorporates specific modifications to enhance its efficacy. A pivotal aspect of these modifications is the selection of the best-scored responses for constructing the alignment data, which forms the basis for the initial phase of supervised fine-tuning. Furthermore, our approach involves a comparative analysis using samples selected via various methods. These include selections based on our scoring system, human judgment, and automated selection by GPT-3.5 or GPT-4.

\paragraph{GPT 4 Judgement of Fine-Tuned OPT 1.3B}
To evaluate the effectiveness of the Supervised Fine-Tuned (SFT) model, we employed GPT-4 to adjudicate the comparative quality of responses. Specifically, we assessed whether the responses generated by the model fine-tuned on selected alignment data exhibited improvements over those produced without such fine-tuning. More detailed judgment prompt design is given below.

    \begin{mdframed}
{\small
    For the following query to the chatbot, which response is more helpful, truthful and harmless to human's question? 

    You need to choose only one of the three answers, respond by either A, B or C. Please do not answer anything other than A, B and C.

    Human: \{prompt\}
    
    A. Assistant: \{answer a\}
    
    B. Assistant: \{answer b\}
    
    C. Assistant: not sure.
    
    Which one to choose? A or B or C?}
    \end{mdframed}
   In our analysis, we classify the comparative assessments into three distinct categories:
    \squishlist
        \item Pre-train: This category is defined by instances where the response with the pre-trained model is deemed superior to the response generated by the supervised fine-tuned model, irrespective of their sequence in presentation.
        \item SFT: the supervised fine-tuned (SFT) model response is consistently evaluated as more effective than the response by the pre-trained model, regardless of their order of appearance.
        \item We remove the cases where the GPT 4 is not sure or inconsistent.
    \squishend
    We report the percentages of SFT being better as evaluated by GPT 4.

\subsection{Experiment Details of \method-Guided In Context Learning (ICL)} \label{app:icl}

\paragraph{Prompt Design of ICL Question-Answering}
We attach the prompt used for ICL question-answering task prepared to query GPT 3.5 as below.
\begin{mdframed}
{\small This task involves answering questions accurately and appropriately. The answers should be concise, respectful, and suitable for a general audience. Below are examples that demonstrate the expected format and content style.
   
    Example Question 1: \{prompt 1\}; 
    
    Answer 1: \{answer 1\}

    ...

    Example Question 5: \{prompt 5\}; 
    
    Answer 5: \{answer 5\}
    
    New Question: \{test question\}; 
    
    Answer: [Your answer here]}
    \end{mdframed}

\paragraph{Comparisons of Text Generation Results}

We attach part of the generation results w.r.t. a single GPT 3.5 and \method ~selected ICL-based prompts. We enumerate questions and their corresponding responses in scenarios where one method demonstrates superior performance according to human verification. This comparative presentation highlights the distinct advantages and potential improvements offered by each approach in different contexts. Generally speaking, \method ~selected samples for ICL-based prompt yields higher quality text generation.

\begin{tcolorbox}[colframe=black,
colbacktitle=gray,title=Question 1: which is the fastest running animal in the world?]

\textbf{Non-Hallucinated Answer:}

the cheetah. The fastest animal in the sea is the black marlin, which has a recorded speed of 129 km/h (80 mph).

  \tcblower
\faThumbsODown ~\textbf{Baseline:} 

The cheetah is the fastest-running animal in the world.
 
\faThumbsOUp ~ \textbf{\method:} 

The cheetah. It is known for its incredible speed and agility, reaching speeds of up to 80 miles per hour (129 kilometers per hour).
\end{tcolorbox}

\begin{tcolorbox}[colframe=black,
colbacktitle=gray,title=Question 2: who sings i must be doing something right?]

\textbf{Non-Hallucinated Answer:}

Billy Currington. Billboard Hot Country Songs chart.

  \tcblower
\faThumbsODown ~\textbf{Baseline:} 

 Billy Currington. The song ``Must Be Doin' Somethin' Right" is sung by Billy Currington.
 
\faThumbsOUp ~ \textbf{\method:} 

Billy Currington. ``Must Be Doin' Somethin' Right" is a song written by Marty Dodson and Patrick Jason Matthews, and recorded by American country music artist Billy Currington.
\end{tcolorbox}

\begin{tcolorbox}[colframe=black,
colbacktitle=gray,title=Question 3: what is the meaning of name iram in urdu?]

\textbf{Non-Hallucinated Answer:}

Heaven. Erum is a Muslim name meaning ``Jannat '' in Persian and Urdu which means `Heaven' in English, with different spellings such as Irum and Iram.
  \tcblower
\faThumbsODown ~\textbf{Baseline:} 

 The name "Iram" does not have a specific meaning in Urdu.
 
\faThumbsOUp ~ \textbf{\method:} 

he meaning of the name Iram in Urdu is ``paradise" or ``garden".
\end{tcolorbox}

\begin{tcolorbox}[colframe=black,
colbacktitle=gray,title=Question 4:  origin of mother's day in the u.s?]

\textbf{Non-Hallucinated Answer:}

the initiative of Ann Reeves Jarvis in the early 20th century. This is not (directly) related to the many traditional celebrations of mothers and motherhood that have existed throughout the world over thousands of years, such as the Greek cult to Cybele, the Roman festival of Hilaria, or the Christian Mothering Sunday celebration ( originally a commemoration of Mother Church, not motherhood ).
  \tcblower
\faThumbsODown ~\textbf{Baseline:} 

 Mother's Day in the United States has its origins in the efforts of Anna Jarvis, who wanted to honor her mother and all mothers for their contributions. She organized the first official Mother's Day celebration in 1908, and it became a recognized holiday in 1914.
 
\faThumbsOUp ~ \textbf{\method:} 

Mother's Day in the U.S. originated from the efforts of Anna Jarvis. She wanted to honor her own mother and all mothers for their contributions to society. In 1908, the first official Mother's Day was celebrated in West Virginia, and it became a national holiday in 1914.
\end{tcolorbox}

\begin{tcolorbox}[colframe=black,
colbacktitle=gray,title=Question 5: who is the girl on fox and friends?]

\textbf{Non-Hallucinated Answer:}

Ainsley Earhardt. Fox \& Friends is a daily morning conservative news/talk program that airs on Fox News Channel, hosted by Steve Doocy, Brian Kilmeade, and Ainsley Earhardt.

  \tcblower
\faThumbsODown ~\textbf{Baseline:} 

Ainsley Earhardt. She joined the show in 2007 and is currently one of the co-hosts of ``Fox \& Friends."
 
\faThumbsOUp ~ \textbf{\method:} 

Ainsley Earhardt. She is one of the co-hosts on the show ``Fox \& Friends."
\end{tcolorbox}

\begin{tcolorbox}[colframe=black,
colbacktitle=gray,title=Question 6: in greek mythology who solved the riddle of the sphinx?]

\textbf{Non-Hallucinated Answer:}

Oedipus. Laius wished to thwart a prophecy , so he left Oedipus to die on a mountainside .

  \tcblower
\faThumbsODown ~\textbf{Baseline:} 

 Oedipus.
 
\faThumbsOUp ~ \textbf{\method:} 

Oedipus. Oedipus solved the riddle of the Sphinx, which allowed him to become the king of Thebes.
\end{tcolorbox}

\begin{tcolorbox}[colframe=black,
colbacktitle=gray,title=Question 7: what book comes after glass in the crank series?]

\textbf{Non-Hallucinated Answer:}

Fallout. Ellen Hopkins 's book, Tilt, was released September 11, 2012, and is a companion from the point of view of the teens mentioned in Triangles.
  \tcblower
\faThumbsODown ~\textbf{Baseline:} 

 Fallout.
 
\faThumbsOUp ~ \textbf{\method:} 

 Fallout. It is the final book in the Crank series written by Ellen Hopkins.
\end{tcolorbox}

\begin{tcolorbox}[colframe=black,
colbacktitle=gray,title=Question 8: who was presided over the first meeting of constituent assembly on 9th december 1946?]

\textbf{Non-Hallucinated Answer:}

Sachchidananda Sinha. Sachchidananda Sinha was elected temporary president of the assembly , in accordance with French practice .
  \tcblower
\faThumbsODown ~\textbf{Baseline:} 

  Dr. Rajendra Prasad.

\faThumbsOUp ~ \textbf{\method:} 

Sachchidananda Sinha.

\end{tcolorbox}

\begin{tcolorbox}[colframe=black,
colbacktitle=gray,title=Question 9:  who was the ventriloquist who won america's got talent?]

\textbf{Non-Hallucinated Answer:}

Terry Wayne Fator. The following year, he was signed on as the headliner at The Mirage hotel and casino in Las Vegas, Nevada, with a five-year, \$100 million contract.
  \tcblower
\faThumbsODown ~\textbf{Baseline:} 

 Darci Lynne Farmer.

\faThumbsOUp ~ \textbf{\method:} 

Terry Fator. He won America's Got Talent in 2007 with his ventriloquism act.
\end{tcolorbox}

\begin{tcolorbox}[colframe=black,
colbacktitle=gray,title=Question 10: where was the first mcdonald's opened outside of the us?]

\textbf{Non-Hallucinated Answer:}

British Columbia, Canada. McDonald's opened its first foreign restaurant in British Columbia, Canada, in 1967.

  \tcblower
\faThumbsOUp ~\textbf{Baseline:} 

 Canada. The first McDonald's outside of the US was opened in Richmond, British Columbia, Canada in 1967.
\faThumbsODown ~ \textbf{\method:} 

 Des Plaines, Illinois. The first McDonald's opened outside of the US was in Des Plaines, Illinois.
\end{tcolorbox}

\section{Limitations and Broader Impacts}\label{app:limits}

\paragraph{Discussion of Limitations.} (1) We still need a certain level of expertise in at least one of the reference LLMs to measure hallucination reasonably accurately. If none of the reference LLMs has any expertise on the question, then the problem is perhaps unsolvable. (2) Our method is slower than merely computing similarity to the gold-standard answers although if we consider the time and resources needed for collecting gold-standard answers, ours is much cheaper.

\paragraph{Impact Statements}
The broader impact of our work revolves around the responsible use of data and maintaining the integrity of large language models (LLMs). Our research aims to measure and reduce LLM hallucination, a major obstacle to the trustworthiness of AI. The method we propose has the potential to influence ethical AI practices in the future. We believe it is essential to thoughtfully address any ethical dilemmas and societal ramifications that may arise from responsible AI. This underscores our dedication to advancing AI technology in a manner that is both responsible and mindful of its broader impacts.

\end{document}